# A Comparative Study of Modern Inference Techniques for Structured Discrete Energy Minimization Problems

Jörg H. Kappes Bjoern Andres Fred A. Hamprecht Christoph Schnörr Sebastian Nowozin
Dhruv Batra Sungwoong Kim Bernhard X. Kausler Thorben Kröger Jan Lellmann
Nikos Komodakis Bogdan Savchynskyy Carsten Rother

April 1, 2014

## Abstract

Szeliski et al. published an influential study in 2006 on energy minimization methods for Markov Random Fields (MRF). This study provided valuable insights in choosing the best optimization technique for certain classes of problems.

While these insights remain generally useful today, the phenomenal success of random field models means that the *kinds* of inference problems that have to be solved changed significantly. Specifically, the models today often include higher order interactions, flexible connectivity structures, large label-spaces of different cardinalities, or learned energy tables. To reflect these changes, we provide a modernized and enlarged study. We present an empirical comparison of 32 state-of-the-art optimization techniques on a corpus of 2,453 energy minimization instances from diverse applications in computer vision. To ensure reproducibility, we evaluate all methods in the OpenGM 2 framework and report extensive results regarding runtime and solution quality. Key insights from our study agree with the results of Szeliski et al. for the types of models they studied. However, on new and challenging types of models our findings disagree and suggest that polyhedral methods and integer programming solvers are competitive in terms of runtime and solution quality over a large range of model types.

## 1 Introduction

Discrete energy minimization problems, in the form of factor graphs, or equivalently Markov or Conditional Random Field models (MRF/CRF) are a mainstay of computer vision research. Their applications are diverse and range from image denoising, segmentation, motion estimation, and stereo, to object recognition and image editing. To give researchers some guidance as to which optimization method is best suited for their MRF models, Szeliski et al. [66] conducted a comparative study on 4-connected MRF models. Along with the study, they provided a unifying software framework that facilitated a fair comparison of optimization techniques. The study was well-received in computer vision community and has till date been cited more than 700 times.

Since 2006 when the study was published, the field has made rapid progress. Modern vision problems involve more complex models, larger datasets and use machine learning techniques to train model parameters.

To summarize, these changes gave rise to challenging energy minimization problems that fundamentally differ from those considered by Szeliski et al. In particular, in [66] the models were restricted to 4-connected grid graphs with unary and pairwise factors only, whereas modern ones include arbitrary structured graphs and higher order potentials.

It is time to revisit the study [66]. We provide a modernized comparison, updating both the problem instances and the inference techniques. Our models are different in the following four aspects:

1. Higher order models, e.g. factors of order up to 300;
2. Models on "regular" graphs with a denser connectivity structure, e.g. 27-pixel neighborhood, or models on "irregular" graphs with spatially non-uniform connectivity structure;
3. Models based on superpixels with smaller number of variables;
4. Image partitioning models without unary terms and unknown number of classes.

Inference methods have changed since 2006 as well, often as a response to the development of challenging models. The study [66] compared the performance of the state of the art at that time, represented by primal move-making methods, loopy belief propagation, a tree-reweighted message passing, and a set of more traditional local optimization heuristics like iterated conditional modes (ICM).

We augment this set with recent updates of the move-making and local optimization methods, methods addressing higher order models, and polyhedral methods considering the energy minimization as an (Integer) Linear Program.

**Contributions** We provide a modernized experimental study of energy minimization methods. Our study includes the cases and algorithms studied by [66], but significantly expand it in the scope of both used inference methods and considered models. Both the methods and the considered models are implemented and stored within a single uniform multi-platform software framework, OpenGM 2 [4]. Together with results of our evaluation they are available on-line on the project webpage [4].

Such a unification provides researchers with an easy access to the considered wide spectrum of modern inference techniques. Our study suggests which techniques are suited for which models. The unification boosts development of novel inference methods by providing a set of models for their comprehensive evaluation.

**Related Inference Studies** Apart from the study [66], a number of recent articles in CV have compared inference techniques for a small specialized class of models, such as [3, 45, 43]. Unfortunately, the models and/or inference techniques are often not publicly available. Even if they were available, the lack of a flexible software-framework which includes these models and optimization techniques makes a fair comparison difficult. Closely related is the smaller study [8] that uses the first and now deprecated version of OpenGM 2. It compares several variants of message passing and move making algorithms for higher order models on mainly synthetic models. In contrast to [8], we consider a broader range of inference methods, including polyhedral ones, and a bigger number of non-synthetic models.

Outside computer vision, the Probabilistic Inference Challenge (PIC) [20] covers a broad class of models used in machine learning. We include the leading optimization techniques and some challenging problems of PIC in our study.

**Key Insights and Suggested Future Research** In comparison with [66], perhaps the most important new insight is that recent, advanced polyhedral LP and ILP solvers are competitive for a wide range of problems in computer vision. For a considerable number of instances, they are able to achieve global optimality. For some problems they are even superior or on a par with approximative methods in terms of overall runtime. This is true for problems with a small number of labels, variables and factors of low order that have a simple form. But even for some problems with a large number of variables or complex factor form, specialized ILP and LP solvers can be applied successfully and provide globally optimal solutions. For problems with many variables and cases in which the LP relaxation is not tight, polyhedral methods are often not competitive. In this regime, primal move-making methods typically achieve the best results, which is consistent with the findings of [66].

Our new insights suggest two major areas for future research. Firstly, in order to capitalize on existing ILP solvers, small but expressive models, e.g. superpixels, coarse-to-fine approaches, or reduction of the model size by partial optimality, should be explored and employed. Secondly, our findings suggest that improving the efficiency and applicability of ILP and LP solvers should and will remain an ongoing active area of research.

A more detailed synopsis and discussion of our insights will be given in Section 7.

The present study is solely devoted to MRF-based energy minimization, that is to MAP estimation from a Bayesian viewpoint. From a statistical viewpoint, inference methods that explore posterior distributions beyond mere point estimation would be preferable, but are too expensive for most large-scale applications of current research in computer vision. Recent work [58, 55] exploits *multiple MAP inference* in order to get closer to this objective in a computationally feasible way. This development underlines too the importance of research on energy minimization as assessed in this paper.

# 2 Graphical Models

We assume that our discrete energy minimization problem is specified on a factor graph $G = (V,F,E)$, a bipartite graph with finite sets of variable nodes $V$ and factors $F$, and a set of edges $E \subset V \times F$ that defines the relation between those [41, 53]. The variable $x_a$ assigned to the variable node $a \in V$ lives in a discrete label-space $X_a$ and notation $X_A$, $A \subset V$, stands for a Cartesian product $\otimes_{a\in A} X_a$. Each factor $f \in F$ has an associated function $\varphi_f : X_{ne(f)} \to \mathbb{R}$, where $ne(f) := \{v \in V : (v, f) \in E\}$ defines the variable nodes connected to the factor $f$. The functions $\varphi_f$ will also be called *potentials*.

We define the order of a factor by its degree $|ne(f)|$, e.g. pairwise factors have order 2, and the order of a model by the maximal degree among all factors.

The energy function of the discrete labeling problem is then given as

$$J(\mathbf{x}) = \sum_{f\in F} \varphi_f(\mathbf{x}_{ne(f)}), \qquad (1)$$

where the assignment of the variable $\mathbf{x}$ is also known as the labeling. For many applications the aim is to find a labeling with minimal energy, i.e. $\hat{\mathbf{x}} \in \arg\min_{\mathbf{x}} J(\mathbf{x})$. This labeling is a maximum-a-posteriori (MAP) solution of a Gibbs distribution $p(\mathbf{x}) = \exp\{-J(\mathbf{x})\}/Z$ defined by the energy $J(\mathbf{x})$. Here, $Z$ normalizes the distribution.

It is worth to note that we use factor graph models instead of Markov Random Field models (MRFs), also known as undirected graphical models. The reason is that factor graphs represent the structure of the underlying problem in a more precise and explicit way than MRFs can, cf. [41].

## 2.1 Categorization of Models

One of the main attributes we use for our categorization is the *meaning* of a variable, i.e. if the variable is associated with a pixel, superpixel or something else. The *number* of variables is typically related to this categorization.

Another modeling aspect is the *number* of labels the variable can take. Note that the size of the label-space restricts the number of methods that are applicable, e.g. QPBO or MCBC can be used when each variable takes no more than two values. We also classify models by *properties* of the factor graph, e.g. average number of factors per node, mean degree of factors, or structure of the graph, e.g. grid structure. Finally, the *properties/type* of the functions embodied by the factors are of interest, since for some subclasses specialized optimization methods exists, e.g. metric energies [66] or Potts functions [33].

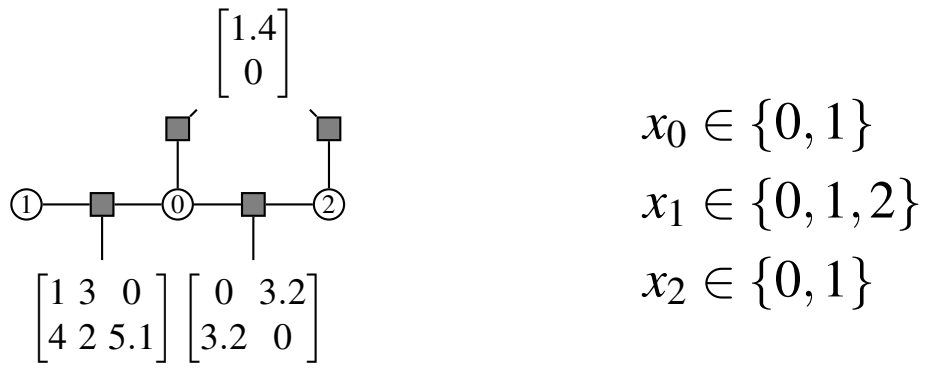

Figure 1: Toy-model constructed by Algorithm 1. The model include 3 variables with 2,3, and 2 labels. The unary factors of variable 0 and 2 share the same function.

## 2.2 OpenGM 2: A General Modeling Framework

For this comparison we use OpenGM 2 [4] a C++ library for discrete graphical models. It provides support for models of arbitrary order, allows the use of arbitrary functions and sharing of functions between factors. OpenGM decouples optimization from the modeling process and allows to store models in the scientific HDF5 format. That is, all model instances stored into a single file and no application specific code has to be released/used to make models available or do evaluation on those. Within OpenGM 2 we provide several own implementations of inference methods in native C++ code as well as wrappers for several existing implementations, and making those available in a unified framework.

Making new model instances or optimization methods available within this framework is rather simple: Fig. 1 illustrates a toy problem with 3 variables having 2,3, and 2 states respectively. The two first order factors represent the same function mapped to different variables. Alg. 1 shows how this is constructed in OpenGM 2. In the first step (lines 1-2) a

**Algorithm 1** Creating a model in OpenGM 2

```
 1: statespace = [ 2 3 2 ]
 2: gm = createModel<+>(statespace)
 3: fid1 = gm.addFunction( [2], [1.4 0] )
 4: fid2 = gm.addFunction( [2 3], [1 3 0; 4 2 5.1] )
 5: fid3 = gm.addFunction( [2 2], Potts(0,3.2) )
 6: gm.addFactor( [0] , fid1 )
 7: gm.addFactor( [2] , fid1 )
 8: gm.addFactor( [0,1] , fid2 )
 9: gm.addFactor( [0,2] , fid3 )
10: storeModel( gm ,"model.h5")
```

graphical model with the variables and corresponding label space is set up. When constructing a model we also fix the operation, in this case addition (+), which couples the single terms to a global objective. Then (line 3-5) the three functions are added. Note that OpenGM 2 allows special implementations for functions, e.g. for Potts functions. In the last step (line 6-9) factors are added to the model and connected to variables (first parameter) and functions (second parameter). Finally the model is stored to file (line 10). OpenGM 2 allows to reuse functions for different factors, which saves a lot of memory if e.g. the same regularizers are used everywhere. We call this concept *extended factor graphs*.

Given a problem defined on an (extended) factor graph one can find the labeling with the (approximately) lowest / highest energy. Alg. 2 illustrates how to approach this within OpenGM 2. After loading a model (line 1), one initializes an

**Algorithm 2** Inference with OpenGM 2

```
1: gm = loadModel("model.h5")
2: InferenceMethod<min,+>::Parameter para
3: InferenceMethod<min,+> inf( gm , para )
4: Visitor vis
5: inf.infer(vis)
6: x = inf.arg()
7: vis.journalize("log.h5")
```

inference object (lines 2-3). Inference is always done with respect to *an accumulative* operation such as minimum (min) or maximum. Optionally, one can set up a visitor object (line 4), which will take care of logging useful information during optimization. The core inference is done within the method "infer" (line 5). After that one can get the inferred labeling (line 6). Additionally, the visitor can give informations about the progress the method has made over time.

For a new inference method, one needs to implement only constructor and methods infer() and arg(). Tutorials for the supported languages can be found on the project website [4].

# 3 Benchmark Models

Table 1 gives an overview of the models summarized in this study. Note that, some models have a single instance, while others have a larger set of instances which allows to derive some statistics. We now give a brief overview of all models. Further specifics in connection with inference will be discussed in Sec. 6. A detailed description of all models is available online.

## 3.1 Pixel-Based Models

For many low-level vision problems it is desirable to make each pixel a variable in the model. A typical property of such models is that the number of variables is large. For 2D images, where variables are associated to pixels in a 2D lattice, a simple form of a factor graph model connects each pixel with its four nearest neighbors (Fig. 2a) using a pairwise energy. This simple form is popular and was the sole subject of the study [66]. In our study we incorporated the models **mrf-stereo**, **mrf-inpainting**, and **mrf-photomontage** from [66] with three, two and two instances, respectively. The pairwise terms of this models are truncated convex functions on the label space for mrf-stereo and mrf-inpainting and a general pairwise term for mrf-photomontage.

Additionally, we used three models which have the same 4-connected structure. For inpainting problems [49] **inpainting-N4** and color segmentation problems [49] **color-seg-N4**[1] the task is to assign each pixel one color out of a preselected finite set. For the object segmentation problems [3] **object-seg** labels correspond to predefined object classes. Each single instance has the same small set of labels for all its variables and Potts terms are used to penalize the boundary length between different classes. In inpainting-N4 and color-seg-N4 this regularizer is the same for all factors. In object-seg, it depends on the image-gradient. The unary terms measure the similarity to predefined class-specific color models.

[1] The inpainting-N4/8 and color-seg-N4/8-models were originally used in variational approaches together with total variation regularizers [49]. A comparison with variational models is beyond the scope of this study.

| | modelname | # | variables | labels | order | structure | functiontype | loss-function | references |
|---|---|---|---|---|---|---|---|---|---|
| **Pixel** | mrf-stereo | 3 | ∼100000 | 16-60 | 2 | grid-N4 | TL1, TL2 | PA2 | [66] |
| | mrf-inpainting | 2 | ∼50000 | 256 | 2 | grid-N4 | TL2 | CE | [66] |
| | mrf-photomontage | 2 | ∼500000 | 5,7 | 2 | grid-N4 | explicit | - | [66] |
| | color-seg-N4/N8 | 2x9 | 76800 | 3,12 | 2 | grid-N4/N8 | potts$^+$ | - | [49] |
| | inpainting-N4/N8 | 2x2 | 14400 | 4 | 2 | grid-N4/N8 | potts$^+$ | - | [49] |
| | object-seg | 5 | 68160 | 4–8 | 2 | grid-N4 | potts$^+$ | PA | [3] |
| | color-seg | 3 | 21000,424720 | 3,4 | 2 | grid-N8 | potts$^+$ | - | [3] |
| | dtf-chinese-char | 100 | ∼ 8000 | 2 | 2 | sparse | explicit | PA | [54] |
| | brain-3/5/9mm | 3x4 | 400000-2000000 | 5 | 2 | grid-3D-N6 | potts$^+$ | - | [17] |
| | inclusion | 1 | 1024 | 4 | **4** | grid-N4 + X | g-potts | PA | [34] |
| **Superpixel** | scene-decomp | 715 | ∼ 300 | 8 | 2 | sparse | explicit | PA | [26] |
| | geo-surf-seg-3 | 300 | ∼ 1000 | 3 | **3** | sparse | explicit | PA | [23, 28] |
| | geo-surf-seg-7 | 300 | ∼ 1000 | 7 | **3** | sparse | explicit | PA | [23, 28] |
| **Partition** | correlation-clustering* | 715 | ∼ 300 | ∼300 | **∼300** | sparse | potts$^*$ | VOI | [38] |
| | image-seg* | 100 | 500-3000 | 500-3000 | 2 | sparse | potts$^*$ | VOI | [6] |
| | image-seg-o3* | 100 | 500-3000 | 500-3000 | **3** | sparse | potts$^*$ | VOI | [6] |
| | knott-3d-seg-150/300/450* | 3x8 | ∼ 800,5000,16000 | ∼ 800–16000 | 2 | sparse | potts$^*$ | VOI | [10] |
| | modularity-clustering* | 6 | 34-115 | 34-115 | 2 | full | potts$^*$ | - | [16] |
| **Other** | matching | 4 | ∼ 20 | ∼20 | 2 | full or sparse | explicit | MPE | [45] |
| | cell-tracking | 1 | 41134 | 2 | **9** | sparse | explicit | - | [36] |
| | protein-folding | 21 | 33-1972 | 81-503 | 2 | full | explicit | - | [72, 20] |
| | protein-prediction | 8 | 14258-14441 | 2 | **3** | sparse | explicit | - | [30, 20] |

Table 1: List of datasets used in the benchmark. Listed properties are number of instances (#), variables and labels, the order and the underlying structure. Furthermore, special properties of the functions are listed; truncated linear and squared (TL1/2), Potts function with positive (potts$^+$) and arbitrary (potts$^*$) coupling strength, its generalization to higher order (g-potts) and functions without special properties (explicit). For some models we have an additional loss function commonly used for this application, namely: 2pixel-accuracy (PA2), color-error (C-E), pixel-accuracy (PA), variation of information (VOI), and geometric error (MPE). For the other models no ground truth or loss function was available.

From a modeling point the 4-neighborhood is quite restricted, important relations cannot be modeled by a simple grid structure in many applications. Therefore, models with denser structures (Fig. 2b) as well as higher order models (Fig. 2c) have been introduced in the last decade. For instance, better approximations of the boundary regularization were obtained by increasing the neighborhood [14]. The datasets **inpainting-N8** and **color-seg-N8** [49] include the same data-term as inpainting-N4 and color-seg-N4 but approximate the boundary length using an 8-neighborhood (Fig. 2b). Another dataset with an 8-neighborhood and Potts terms depending on the image-gradient is **color-seg** by Alahari et al. [3].

We also use a model with a 6-neighborhood connectivity structure in a 3D-grid. It is based on simulated 3D MRI-brain data [17], where each of the 5 labels represent color modes of the underlying histogram and boundary length regularization [14]. We let the simulator generate 4 scans for 3 different slice-thickness. These models are denoted by **brain-9mm**, **brain-5mm**, and **brain-3mm**. These replace the brain dataset used in [31].

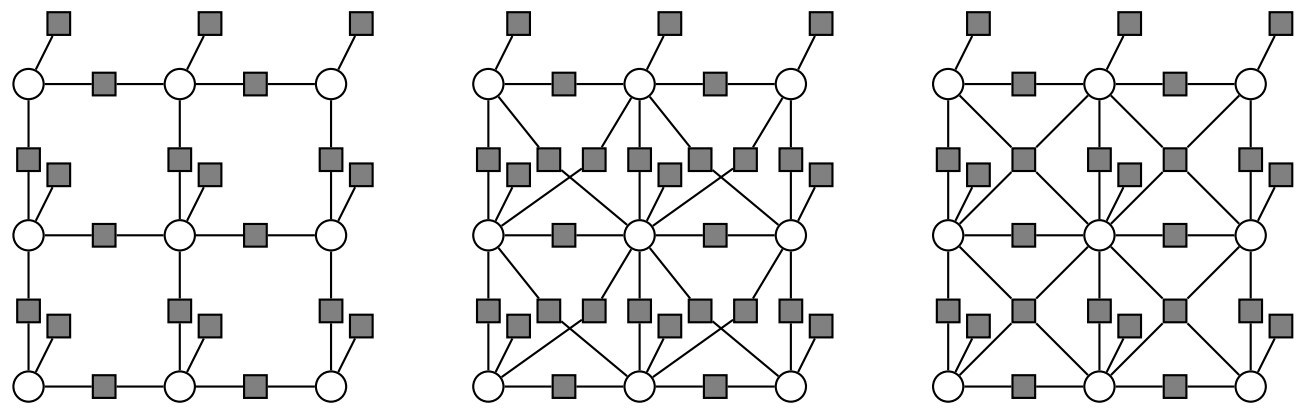

(a) Grid N4 (b) Grid N8 (c) Grid HO

Figure 2: Common **pixel based models** are grid structured with respect to a four (a) or eight (b) neighborhood-structure. Some models also use couplings to remoter variables. Also higher order structures (c) have been successfully used for modeling.

We also consider the task of inpainting in binary images of Chinese characters, **dtf-chinesechar** [54]. Potentials, related to the factors of these models, are learned from a decision tree field. Although each variable has only two labels, it is connected via pairwise factors to 27 other variables selected during learning from a $17 \times 17$ window. Such an increased connectivity and discriminative learned potential make the resulting inference problem highly non-sub-modular and therefore challenging.

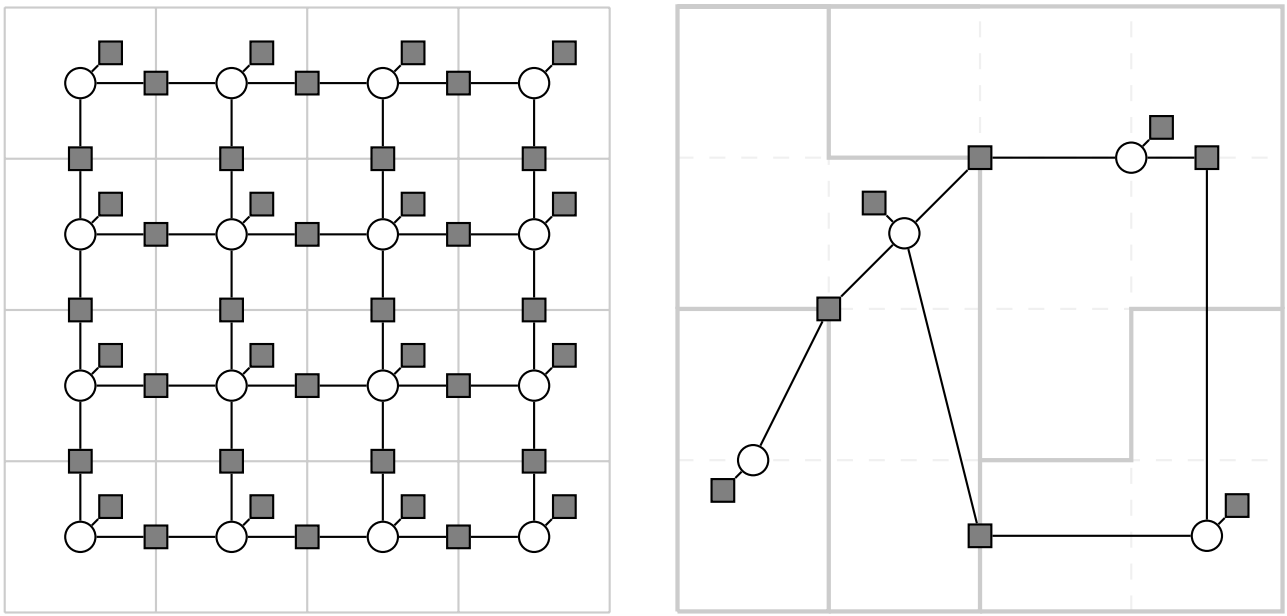

(a) pixel based (b) superpixel based

Figure 3: While in pixel based models (a) each variable/node is assigned to one pixel, in superpixel based models (b) each variable/node is assigned to a set of pixels forming the superpixel. As a consequence the number of variables becomes smaller and the graph-structure is usually no longer regular.

Pixel based models that include higher-order terms are really rare. The main reason for this is that pixel based models usually have a large number of variables, such that a systematical enrichment of the model with higher-order terms often becomes intractable. To cope with this, higher order models in computer vision often include terms that can be reduced linearly to a second order model by including a small number of auxiliary variables and additional factors, e.g. labeling costs [19] and $P^n$-Potts [40] terms. Our benchmark includes one model called **inclusion** that has a fourth-order term (Fig. 2c) for each junction of four pixels that penalizes a splitting of the boundary, as suggested in [34].

## 3.2 Superpixel-Based Models

In superpixel-based models, all pixels belonging to the same superpixel are constrained to have the same label, as shown in Fig. 3. This reduces the number of variables in the model and allows for efficient inference even with more complex, higher order factors. However, the grouping of pixels is an irreversible decision and it is hard to treat grouped pixels differently later on.

The models we consider in this regime are used for semantic image segmentation. The number of superpixels vary between 29 and 1133 between the instances. In the **scene-decomposition**-dataset [26] every superpixel has to be assigned to one of 8 scene classes. Pairwise factors between neighboring superpixels penalize unlikely label-pairs. The datasets **geo-surf-3** and **geo-surf-7** [23, 28] are similar but have additional third-order factors that enforce consistency of labels for three vertically neighboring superpixels.

## 3.3 Partition Models

Beyond classical superpixel models, this study also considers a recent class of superpixel models which aim at partitioning an image without any class-specific knowledge [38, 6, 9, 10]. These use only similarity measures between neighbored regions encoded by (generalized) Potts functions with positive and negative coupling strength. Since the partition into isolated superpixels is a feasible solution, the label space of each variable is equal to the number of variables of the model, and therefore typically very large, cf. Tab. 1.

For unsupervised image segmentation we consider the probabilistic image partition dataset **image-seg** [6], which contains factors between pairs of superpixels, and its extension **image-seg-o3** [6] that also uses a learned third-order prior on junctions. The 3d neuron segmentation model 3d-neuron-seg used in [31] are replaced by 3 datasets **knott-3d-seg-150**, **knott-3d-seg-300**, and **knott-3d-seg-450** with 8 instances each. The number denotes the edge length of the 3D volume. These datasets allow a better evaluation of the scalability of methods and are generated from sub-volumes of the model described in [9, 10].

The hyper-graph image segmentation dataset **correlation-clustering** [38] includes higher order terms that favor equal labels for sets of superpixels in their scope if those are visually similar. These sets are pre-selected and incorporate higher level proposals in the objective. The partition models are completed by some network clustering problems **modularity-clustering** [16] from outside the computer vision community. Contrary to the previous ones, these instances include a fully connected structure.

## 3.4 Other Models

We also consider computer vision applications that assign variables to keypoints in the image-data.

The first model deals with the non-rigid point matching problem [45] **matching**. Given two sets of keypoints the task is to match these such that the geometric relations are kept. The model instances include no unary terms, whereas the pairwise terms penalize the geometric distortion between pairs of points in both point-sets.

The second application is **cell-tracking** [36]. Variables correspond to the assignments of cells in a video sequence, which need to be consistent over time. Since a track can either be active or dormant, the variables are binary. Higher-order factors are used to model the likelihood of a ”splitting” and ”dying” event of a cell.

Finally, we include models from outside computer vision, taken from the Probabilistic Inference Challenge (PIC) [20] into the corpus. The **protein-folding** instances [72] have a moderate number of variables, but are fully connected and have for some variables huge label spaces. The **protein-prediction** instances [30] include sparse third-order binary models. For both dataset we include only the hardest ones, which were used also in [20].

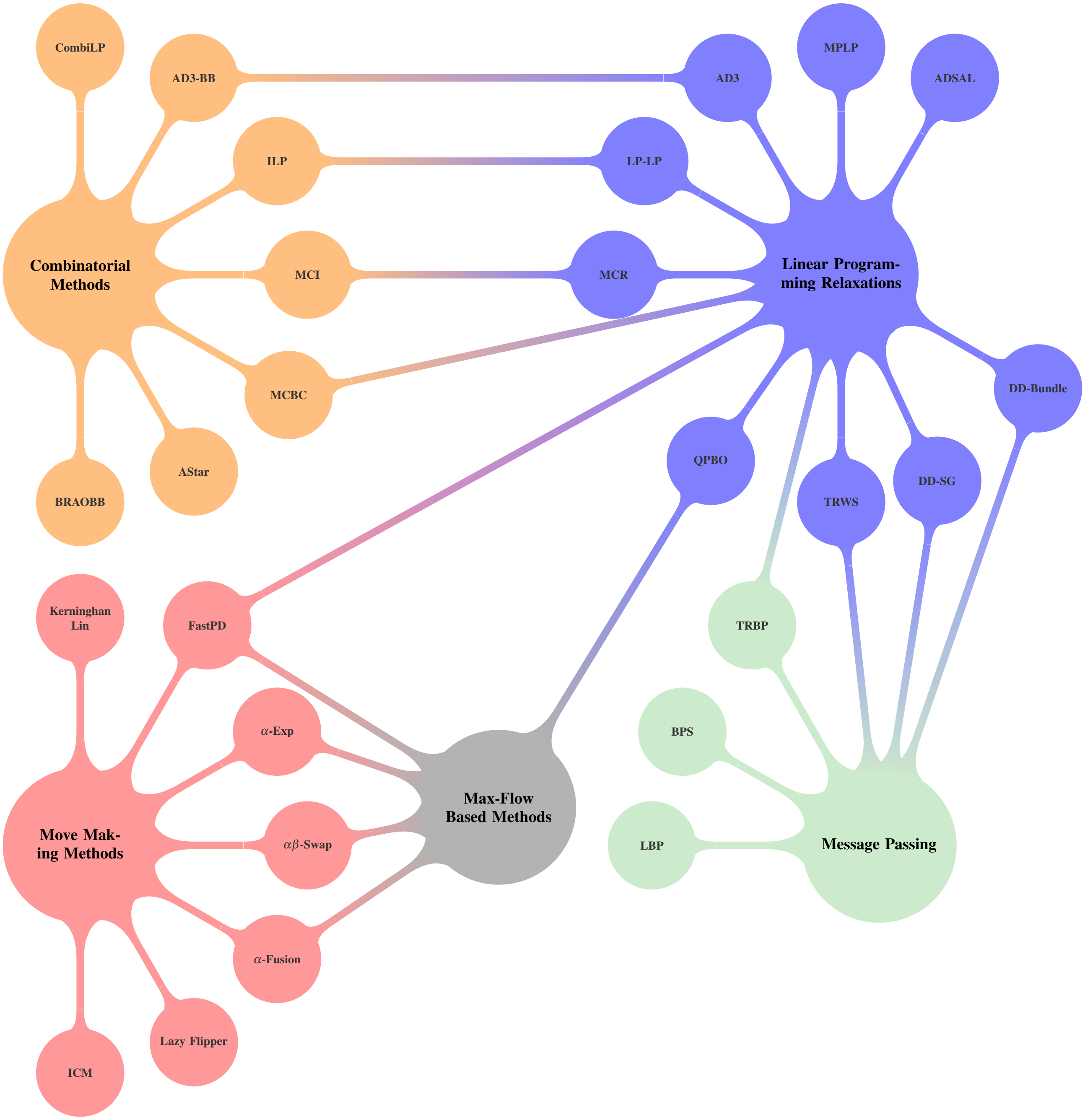

Figure 4: The inference methods used in this benchmark can be roughly grouped into four classes. These are (1) methods based on **linear programming**, (2) methods based on **combinatorial optimization**, which are often strongly related to linear programming, (3) methods based on **move-making** procedures which iteratively improves the labeling, and (4) methods based on **message passing** – often motivated by linear programming and variational optimization. Some methods make use of max-flow methods for fast optimization of binary (sub-)problems, which is also sketched in the diagram. A fifth class of methods are methods based on **sampling**, which are not covered in this study since they are rarely used in computer vision. For hard models they might perform reasonable, with a certain amount of tuning of involved hyper-parameters and sampling procedures, as shown for the dtf-chinesechar model.
For some of the inference algorithms we use different implementations. Even when algorithmically identical, they often vary in speed because of implementation differences and specialized algorithms. We always try to use the fastest one and use the prefix ogm- and mrf- to state that the used implementation was [5] or [66], respectively. For other methods the core of the implementation has been provided by the original authors of the methods and we wrapped them within OpenGM 2.

# 4 Inference Methods

We evaluate a large number of different inference methods. The selection of methods is representative of the state of the art in the field. As shown in Fig. 4, we can cluster the methods into four groups: (i) combinatorial methods that provide optimal solution if no time-constraints are given, (ii) methods solving linear programming relaxations, providing a lower bound, (iii) move-making methods that iteratively improve a feasible labeling, and (iv) message passing methods that perform local calculations between nodes. Furthermore, a subset of methods leverage max-flow problems; we call these methods max-flow-based methods. In this study we do not consider inference algorithms based on Monte Carlo simulation; only in one data set (dtf-chinesechar) we report results obtained from simulated annealing. In general these methods converge slowly and may not be practical for most computer vision applications. In addition they often require careful tuning of parameters such as the temperature schedule. Because Monte Carlo methods are not very popular in computer vision, we are not able to give a fair comparison and skip them in our current study. We now give a brief overview of the considered methods.

## 4.1 Polyhedral Methods

A large class of algorithms solve a linear programming relaxation (LP) of the discrete energy minimization problem. An advantage of these methods is that they also provide a lower bound for the optimum, but on the other hand can converge to non-integer solutions, which need to be rounded to an integer solution.

Perhaps the most commonly used relaxation is the LP relaxation over the *local polytope* [69, 64, 70]. We can solve small instances using off-the-shelf LP-solvers e.g. CPLEX [18] as used in **LP-LP** [5]. For large problems this is no longer possible and special solvers have been proposed that optimize a dual formulation of the problem. A famous example is the block-coordinate-ascent method **TRWS** [42], which, however, can get stuck in suboptimal fix points.

In contrast, subgradient methods [46, 32] based on dual decomposition (DD) [46, 27] with adaptive stepsize **DD-SG-A** and bundle methods [32] with adaptive **DD-Bundle-A** or heuristic **DD-Bundle-H** stepsize are guaranteed to converge to the optimum of the relaxed dual[2]. In both cases primal integer solutions are reconstructed from the subgradients. Because dual decomposition methods can be sensitive to the stepsize, we evaluate different stepsize-rules. While a more detailed evaluation is beyond the scope of this works, we consider beside our base line stepsize-rule also stepsizes scaled by 10 and $10^{-1}$, denoted with postfix $+$ and $-$, respectively.

Other methods based on dual decomposition are Alternating Directions Dual Decomposition **AD3** [51], the Adaptive Diminishing Smoothing ALgorithm **ADSAL** [63], which smooth the dual problem to avoid local suboptimal fix-points, and Max-Product Linear Programming **MPLP** [24]. For MPLP an extension **MPLP-C** [65] exists that iteratively adds violated constraints over cycles of length 3 and 4 to the problem. This leads to a tighter relaxations than the local polytope relaxation.

**Algorithm 3** Dual-Decomposition

1: **Decompose problem**: $E(x|\theta) = \sum_i E_i(x^i|\theta^i)\ s.t. x^i \equiv x$
2: **repeat**
3: **Solve subproblems**: $\forall i: \ x^{i*} = \arg\min_{x^i} E_i(x^i|\theta^i + \lambda^i)$
4: **Update dual variables**: $\forall i: \ \lambda^i = \lambda^i + \xi^i(x^{i*}, \ldots, x^{n*})$
5: **Ensure by projection that**: $\sum_i \lambda^i \equiv 0$
6: **until** Stopping condition is fulfilled

The idea of most dual methods is sketched in Alg. 3. Starting with a decomposition of the original into several tractable subproblems, the equality constraints $x^i \equiv x$ are dualized by means of Lagrange multipliers. The subgradients, which can be obtained by solving the subproblems, are used to update the dual variables $\lambda$. If the update has moved $\lambda$ to a point outside the feasible set an additional projection onto the feasible set is required. This step can also be understood as a re-parameterization of the problem, with the goal that the subproblems will agree in their solutions, i.e. $x^i \equiv x^j$. The different dual methods mainly distinguish in the choice of decomposition (Alg. 3 line 1) and the used update strategy $\xi^i$ for dual variables (Alg. 3 line 4).

For binary second order problems the **QPBO** method [60] can be used to find the solution of the local polytope relaxation in low order polynomial time. It reformulates the LP as a network flow problem, which is then solved efficiently. For Potts models we also compare to a cutting-plane algorithm, **MCR** [33], that deals with a polynomially tractable relaxation of the multiway cut polytope and the multicut polytope. For positive coupling strength the former polytope is equivalent to the local polytope relaxation [56, 52]. We compare different types of relaxations and separation procedures as described in [35][3].

## 4.2 Combinatorial Methods

Related to polyhedral methods are Integer Linear Programs (ILPs). These include additional integer constraints and guarantee global optimality, contrary to the methods based on LP-relaxations which may achieve optimality in some cases. Solutions of ILPs are found by solving a sequence of LPs and either adding additional constraints to the polytope (cutting plane techniques) as sketched in Alg. 4, or branching the polytope or discrete candidate-set into several polytopes or candidate-sets (Branch and Bound techniques), sketched in Alg. 5. Inside Branch and Bound methods the cutting plane methods can be applied to get better bounds which allow to exclude subtrees of the branch-tree earlier.
We evaluate four state-of-the-art general combinatorial solvers. We wrapped the off-the-shelf ILP solver from IBM

[2] Here we consider spanning trees as subproblems such that the relaxation is equivalent to the local polytope relaxation.

[3] This includes terminal constraints *TC*, multi-terminal constraints *MTC*, cycle inequalities *CC* and facet defining cycle inequalities *CCFDB* as well as odd-wheel constraints *OWC*.

**Algorithm 4** Cutting-Plane

**Initial Relaxation**: $\min_{\mu \in P} \langle \theta, \mu \rangle$
**repeat**
  **Solve current relaxation**: $\mu^* = \arg\min_{\mu \in P} \langle \theta, \mu \rangle$
  **Add constraints violated by $\mu^*$ to P**
**until** No violated constraints found

**Algorithm 5** Branch and Bound

**Initial Problem**: $\min_{x \in X} E(x)$, $S = \{X\}$
**repeat**
  **Select branch node**: $\tilde{X}$ with $\tilde{X} \in S$
  **Split selected node**: $\tilde{X}^1, \ldots, \tilde{X}^n$ with $\tilde{X} = \tilde{X}^1 \cup \ldots \cup \tilde{X}^n$
  **Branch**: $S = (S \setminus \tilde{X}) \cup \tilde{X}^1 \cup \ldots \cup \tilde{X}^n$
  **Bound**: $\forall i = 1, \ldots, n : B^{\tilde{X}^i} \leq \min_{x \in \tilde{X}^i} E(x)$
  **Solution (if possible)**: $\forall i = 1, \ldots, n : V^{\tilde{X}^i} = \min_{x \in \tilde{X}^i} E(x)$
**until** $\exists s \in S : V^s \leq \min_{s \in S} B^s$

CPLEX [18] by OpenGM 2 [5] and denoted by **ILP**. For the best performing method in the PIC 2011, called breath-rotating and/or branch-and-bound [57] (**BRAOBB**), we observe for several instances that the optimality certificate, returned by the algorithm, is not correct. We reported this to the authors, who confirmed our observations; we chose to not report the invalid bounds returned by BRAOBB. We evaluate three variants: BRAOBB-1 uses simple and fast pre-processing, BRAOBB-2 uses stochastic local search [29] to quickly find a initial solution, and BRAOBB-3 is given more memory and running time to analyse the instances during pre-processing. Bergtholdt et al. [12] suggested a tree-based bounding heuristic for use within A-star search [12]; we call this method **A-Star**. This Branch & Bound method does not scale to large problems. The same is true for the Branch & Bound extension of AD3, which uses upper and lower bounds of AD3 for bounding; we denote this extended method by **AD3-BB** [51].

The recently proposed **CombiLP** solver [61] utilizes the observation that the relaxed LP solution is almost everywhere integral in many practical computer vision problems. It confines application of a combinatorial solver to the typically small part of the graphical model corresponding to the non-integer coordinates of the LP solution. If consistence between the LP and ILP solution cannot be verified the non-integral subparts grow and the procedure repeats. This allows to solve many big problems exactly. If the combinatorial subproblem becomes too large, we return bounds obtained by the LP solver.

To reduce the large memory requirements, we also consider the integer multicut-representation introduced by Kappes et al. [33]. This multicut solver with integer constraints (**MCI**) can only be applied for functions which include terms that are either invariant under label permutations or of the first-order. As for MCR similar separation-procedures are available. Additionally, we can take advantage from integer solutions and use more efficient shortest-path methods, noted by an *I* within the separation acronym.

We also consider a max-cut branch and cut solver **MCBC** for pairwise binary problems, which could not be made publicly available due to license restrictions.

## 4.3 Message-Passing Methods

Message passing methods are simple to implement and can be easily parallelized, making them a popular choice in practice. The basic idea is sketched in Alg. 6. In the simplest case messages are defined on the edges of the factor graph, better approximations can be obtained by using messages between regions. Messages can be understood as a re-parameterization of the model, such that local optimization becomes globally consistent. Polyhedral methods can often be reformulated as message passing where the messages represent the re-parameterization of the models, as in **TRWS** and **MPLP**. Its non-sequential pendant **TRBP** [69] is written as a message passing algorithm. TRBP can be applied to higher order models but has no convergence guarantees. Practically it works well if sufficient message damping [69] is used. Maybe the most popular message passing algorithm is loopy belief propagation (LBP). While LBP converges to the global optimum for acyclic models, it is only a heuristic for general graphs, which turns out to perform reasonably well in practice [73]. We evaluate the parallel (**LBP**) and sequential (**BPS**) versions from [66], as well the general higher order implementation using parallel updates from [5]. For non-sequential methods we use message damping.

**Algorithm 6** Message Passing

**Setup**: $\forall e \in E$ : initialize a message for each direction.
**repeat**
  **Update**: $\forall e \in E$ : update the message given the other messages.
**until** no significant change of messages
**Decoding**: Re-parameterize the original problem by the messages and decode the state locally or greedy.

Another advantage of message passing methods is that they can be easily parallelized and further speeded up for the calculation of the message updates in special cases, e.g. when distance transform [21] is available.

## 4.4 Move-Making Methods

Another class of common greedy methods applies a sequence of minimizations over subsets of the label space, iteratively improving the current labeling. The corresponding subproblems have to be tractable and the current label has to be included into the label-subset over which the optimization is performed, cf. Alg. 7.

The $\alpha$-$\beta$-Swap-Algorithm [15, 44] (**$\alpha$-$\beta$-Swap**) selects two labels, $\alpha$ and $\beta$, and considers moves such that all variables currently labelled $\alpha$ or $\beta$ are allowed to either remain with their current label or change it to the other possible label within the set $\{\alpha, \beta\}$. In each round all possible pairs of labels

**Algorithm 7** Move Making Methods

```
1: Setup: Select an initial labeling x* ∈ X
2: repeat
3:    Select Move: X̃ ⊂ X s.t. x* ∈ X̃
4:    Move/Update: x* = argmin_{x∈X̃} E(x)
5: until No more progress is possible
```

are processed. For each label pair the corresponding auxiliary problems are binary and under some technical conditions submodular, hence they can be solved efficiently by max-flow algorithms, see [68] for a recent review.

Alternatively, the same authors [15, 44] suggested the $\alpha$-Expansion algorithm (**$\alpha$-Exp**). This move making algorithm selects a label $\alpha$ and considers moves which allows all variables to either remain with their current label or to change their label to $\alpha$. In each round $\alpha$ is sequentially assigned to all possible labels. As for $\alpha$-$\beta$-Swap the auxiliary problem in 7 line 3 can be reduced to a max-flow problem under some technical conditions.

The **FastPD** algorithm [47] is similar to $\alpha$-Expansion in the way moves are selected and evaluated. Additionally, the dual solution of the max-flow solver is used to re-parameterize the objective function. This leads to significant speed up and allows as a by-product to calculate a lower bound by using the re-parameterized objective. However, FastPD might get stuck in suboptimal fix-points and will not reach the optimal dual objective.

While these three methods are based on solvable max-flow subproblems, there are other algorithms which perform local moves. The most prominent algorithm in this line is the iterated conditional modes algorithm [13] (**ICM**), which iteratively improves the labeling of a single variable by keeping the other fixed. Andres et al. [7] extend this idea to an exhaustive search for all sets of variables of a given size $k$. While this is intractable in a naive implementation, they introduce several data-structures to keep track of the changes and end up with the **Lazy Flipper**. As with ICM this method also guarantees to converge to a local fix-point that can only be improved by changing more than 1 or $k$ variables for ICM and Lazy Flipper, respectively.

We also consider generalizations of $\alpha$-Expansion for general problems and higher order factors, which is a special case of the class of fusion-methods [50] and therefore called **$\alpha$-Fusion**. As proposal oracles we use labelings with the same label for all variables. While other proposal-oracles will perform better for some models, finding better oracles is application dependent and beyond the scope of this work, but indeed very interesting. The auxiliary problems are solved by QPBO, which provides partial optimality, and variables with non-integer solutions are not changed within the move. In order to deal with higher-order models we apply the reduction techniques of Fix et al. [22] to the auxiliary problems.

Another method specialized for partition problems is the Kernighan-Lin algorithm [37] (**KL**), which iteratively merges and splits regions in order to improve the energy.

## 4.5 Rounding

Linear programming relaxations and message passing methods typically do not provide a feasible integer solution. They provide either a fractional indicator vector or pseudo-min-marginals. The procedure to transform those into a feasible integer solution is called rounding. The final quality of the integer solution does also depend on the used rounding procedure.

The simplest rounding procedure is to round per variable based on the first order indicator function or marginal denoted by $b_a$

$$\forall v \in V: \qquad x_v^* = \mathrm{opt}_{x_v \in X_v} b_v(x_v) \tag{2}$$

While this is simple and can be performed efficiently it is very brittle and might fail even for tree-structured models by mixing two modes. Ideally, decisions should not be made independently for each variable. One popular option is to condition eq. (2) on already rounded variables

$$\forall v \in V: \qquad x_v^* = \mathrm{opt}_{x_v \in X_v} \sum_{\substack{f \in ne(v) \\ \text{and all nodes } ne(f)\setminus\{v\} \text{ are fixed}}} b_f(x_v, x^*_{ne(f)\setminus v}) \tag{3}$$

Different implementations use different rounding procedures, e.g. MCR uses a de-randomization procedure [35] which gives additional guarantees, the implementation of MPLP applies different rounding schemes in parallel and selects the best performing one. However, since the rounding problem can be as hard as the original problem, which is NP-hard, there will always be cases in which one rounding method performs worse than another. It is important to understand that when we compare the objective value of the rounded integer solution of a method we implicitly also evaluate the rounding method used.

## 4.6 Post- and Pre-Processing

By combining mentioned solvers or augmenting them with other procedures, the further "meta-solvers" can be generated. The **CombiLP** is one example for this. It combines TRWS with ILP.

Alahari [2] suggested a pre-processing procedure that first computes the optimal states for a subset of variables in polynomial time and then continues with the remaining smaller problem with more involved solvers. In [2] this was proposed in order to improve the runtime, and [35] showed that using this preprocessing can make combinatorial methods as efficient as approximate methods. In [35] the authors also suggest to treat acyclic substructures and connected components separately. We combine different methods with this preprocessing denoted by the postfix **-ptc**. For Potts models we make use of Kovtun's method [48] to efficiently obtain partially optimal solutions.

An alternative approach to reduce the problem size is to group variables based on the energy function [39]. However, this has two major drawbacks; first it is not invariant to model reparameterizations, and second the reduced problem is an approximation of the original problem.

Another meta-solver we considered is InfAndFlip, which first runs a solver to obtain a good starting point for the Lazy

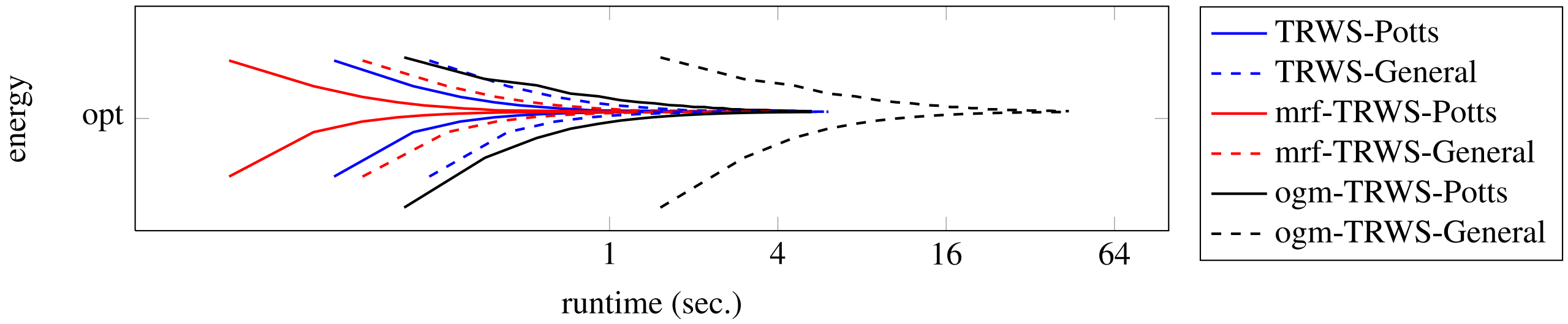

Figure 5: Logarithmically compressed value and bound plots for six different implementations of TRWS for the same instance. Using distance transform for Potts models always leads to a speed up compared to standard updates. The mrf-TRWS which is specialized for grids is fastest, followed by the original implementation TRWS which requires the same regularizer type everywhere, and ogm-TRWS which only restricts the model to be of second order.

Flipper. Lazy Flipping never degrades the energy, but relies heavily on its initialization. We denote lazy flipping post-processing by **-LFk**, where *k* specifies the search depth.

## 4.7 Comparability of Algorithms and their Implementations

When we empirically compare algorithms, we really compare specific *implementations* of said algorithms, which are effected by domain-specific implementation choices. It is important to keep in mind that a general implementation will usually be much slower than a specialized one. The reason is that specialized efficient data structures can be used, more efficient calculations for subproblems become feasible, interfaces can be simplified, compiler optimization (e.g. loop unrolling) becomes applicable and many more or less obvious side-effects can show up.

Our study includes implementations that follow two different paradigms; very fast, specialized but less flexible code, e.g. TRWS and BPS of Vladimir Kolmogorov or FastPD of Nikos Komodakis, and very general and flexible but less fast code, e.g. the OpenGM 2 implementations of Dual Decomposition or LBP.

While both paradigms have merit, it becomes challenging to quantify their relative performance in a fair manner. Due to the algorithmic complexity we expect that for some methods a speedup of a factor of $\approx 100$ for specialized implementations may be possible.

Specialized solvers are very fast for problem classes they support and were designed for, but often are not easily generalized to other classes and at that point restrict the degrees of freedom of modeling. On the other hand more general solvers are able to solve a large class of models, but are often orders of magnitudes slower. Specifically, some solvers are specialized implementations for a certain class of problems (e.g. grids with Potts functions), while others make no assumptions about the problem and tackle the general case (i.e. arbitrary functions and order).

As one of several possible examples Fig. 5 shows six different implementations of TRWS. The original TRWS implementation for general graphs (TRWS) and grid graphs (mrf-TRWS) by Vladimir Kolmogorov and the TRWS implementation provided in OpenGM 2 (ogm-TRWS). Each is implemented for general functions and for Potts functions using distance transform [21].

## 4.8 Algorithmic Scheduling

A number of algorithms depend on an explicit or implicit ordering of factors, messages, labels, etc. While the detailed evaluation of such implementation details or implicit parameters is beyond the scope of this work, we describe the choices in our implementations and possible alternatives for the sake of completeness.

For $\alpha$-Expansion and $\alpha\beta$-Swap we can chose the default order of the moves. Recently, Batra and Kohli [11] have shown, that an optimized ordering can often improves the results. However, there are no theoretical guarantees that this algorithm will produce an ordering strictly better than the default one. Similarly, the order of moves has a strong impact on the quality of solutions and runtime in the case of other move-making algorithms such as ICM and Lazy-Flipper.

Sequential message passing methods depend on the ordering of the messages. This ordering can be predefined or dynamically selected during optimization, as suggested by Tarlow et al. [67]. Parallel message passing methods do not require an ordering, but typically underperform compared to asynchronous algorithms and often require damping to achieve convergence. While empirically sequential methods perform better, they are not as easy to parallelize.

For methods using cutting plane techniques and branch and bound, the cutting plane management/scheduling [71] and branching strategy [1] define an additional degree of freedom, respectively. We always use the default option of the methods which provide in average good choices.

For our experiments we use the alphabetical ordering on the model or the ordering used in external code by default.

## 4.9 Stopping Condition

Another critical point which has to be taken into account is the stopping condition used. Not all methods have a unique stopping condition. For example move making methods stop if no move from the considered set of moves gives an improvement. This is practical only if the set of possible moves is manageable, for example if it can be optimized over exactly.

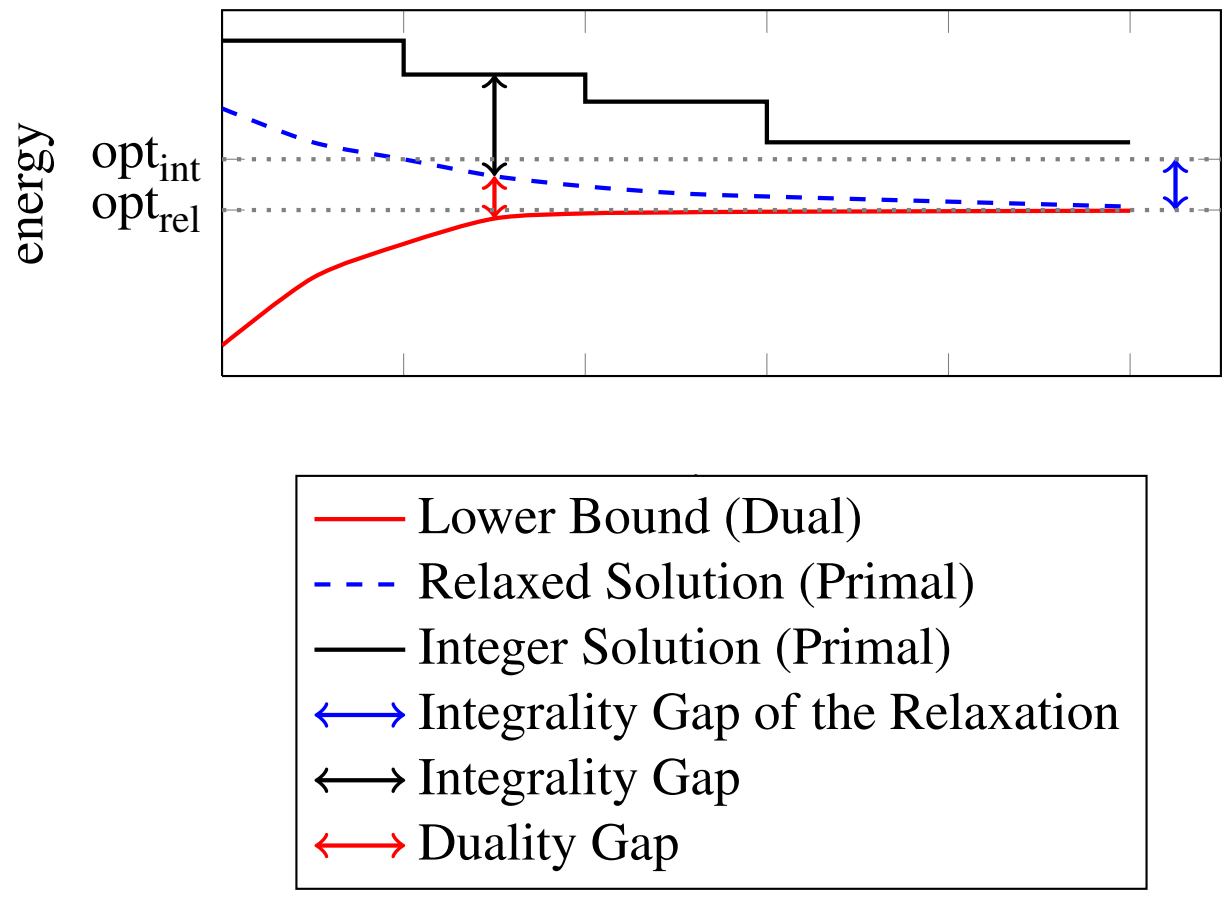

Figure 6: Illustration of temporal changes of the objectives within linear programming frameworks. LP solves either the relaxed problem (blue) or its dual problem (red). Under mild technical condition their optimal value is identical, i.e. the duality gap vanishes. Extracting a primal solution from a dual is non-trivial and typically the solution is non-integral. Rounding to a feasible integer solution causes an additional integrality gap.

In order to deal with the accuracy of floating-point arithmetic [25], linear programming approaches often solve the problem up to a certain precision, here set to $10^{-7}$. Solving the LP to a higher precision requires more runtime, but may not improve (or even change) the final integer solution since LPs typically involve a rounding procedure.

For LP solvers that work on the dual problem like TRWS or MPLP it is non-trivial to evaluate the duality gap. To overcome this problem Savchynskyy and Schmidt [62] propose a method to generate reasonable feasible primal estimates from duals and so get an estimate on the primal-dual-gap.

Unfortunately, in theory even small changes in the primal-dual-gap can have large impact on the integrality gap. Moreover, the method of Savchynskyy and Schmidt is so far not available for all algorithms, so we use the total gap (integrality + duality gap) as our stopping condition, cf. Fig. 6.

Another option is to check if the changes made by the method are numerically relevant. If for example the largest change of a message of LBP is smaller than $10^{-5}$ it is more likely to run into numerical problems than to make further improvements. The same holds true if the improvements of the dual within TRWS become too small.

We will use the following additional stopping conditions for all algorithms: **(1)** A method is stopped after 1 hour. **(2)** A method is stopped if the gap between the energy of the current integer solution and the lower bound is smaller than $10^{-5}$. **(3)** A method is stopped if the numerical changes within its data is smaller than $10^{-7}$.

With this additional stopping condition, we obtain better numerical runtimes for some methods, e.g. TRWS and LBP, as reported in [31] without worsening the other results. Such methods suffer when a large number of iterations is used as the only stopping-condition.

# 5 Experimental Setup

The hardware platform for this study is the Intel Core i7-2600K 3.40GHz CPU, equipped with 16 GB of RAM[4]. In order to minimize the effect of operating system activity on runtime measurements, experiments were conducted on only three of the four cores of a CPU. No multi-threading and no hyper-threading was used. An evaluation of the parallelizability of optimization algorithms or the runtime of parallel implementations is beyond the scope of this study.

The software platform for this study is Version 2.3.1 of the C++ template library OpenGM [4, 5]. OpenGM imposes no restrictions on the graph or functions of a graphical model and provides state-of-the-art optimization algorithms, custom implementations as well as wrappers around publicly available code. In the tables below, prefixes indicate the origin of each particular implementation, *ogm* [5] and *mrf* [66]. The lack of a prefix indicates that code was provided by the corresponding author and wrapped for use in OpenGM. All graphical models considered in this study are available from [4] in the OpenGM file format, a platform independent binary file format built on top of the scientific computing standard HDF5.

To make runtime measurements comparable, we exclude from these measurements the time for copying a graphical model from the OpenGM data structure into a data structures used in wrapped code, the time spent on memory allocation in the initialization phase of algorithms, as well as the overhead we introduce in order to keep track, during optimization, of the current best integer solution and, where available, the current best lower bound. To keep resources in check, every algorithm was stopped after one hour if it had not converged.

Obviously, not every implementation of every algorithm is applicable to all graphical models. We made our best effort to apply as many algorithms as possible to every model. As a consequence of this effort, the study compares implementations of algorithms which are highly optimized for and restricted to certain classes of graphical models with less optimized research code applicable to a wide range of graphical models. As discussed in Section 4.7, this aspect needs to be taken into account when comparing runtimes.

# 6 Evaluation

This section summarizes the experimental results. A compact overview of the performance of the algorithms is given in Fig. 14–22. Instructions for reading these figures are given in Fig. 13. For selected graphical models and instances, proper numbers are reported in Tables 2–22. For all models and instances, proper numbers as well as graphs of upper and lower bounds versus runtime can be found online [4].

All tables in this section as well as all tables online show the *runtime*, the objective *value* of the final integer solution as well as the lower *bound*, averaged over all instances of a particular model. In addition, we report the number of instances for which an algorithm returned the *best* (not necessary optimal) integer solution among all algorithms and the number of instances for which the algorithm verified optimality, denoted by *opt*[5].

[4]Due to the increased workload compared to the experiments in [31], we switch to a homogeneous cluster and no longer use the Intel Xeon W3550 3.07GHz CPU equipped with 12 GB RAM.

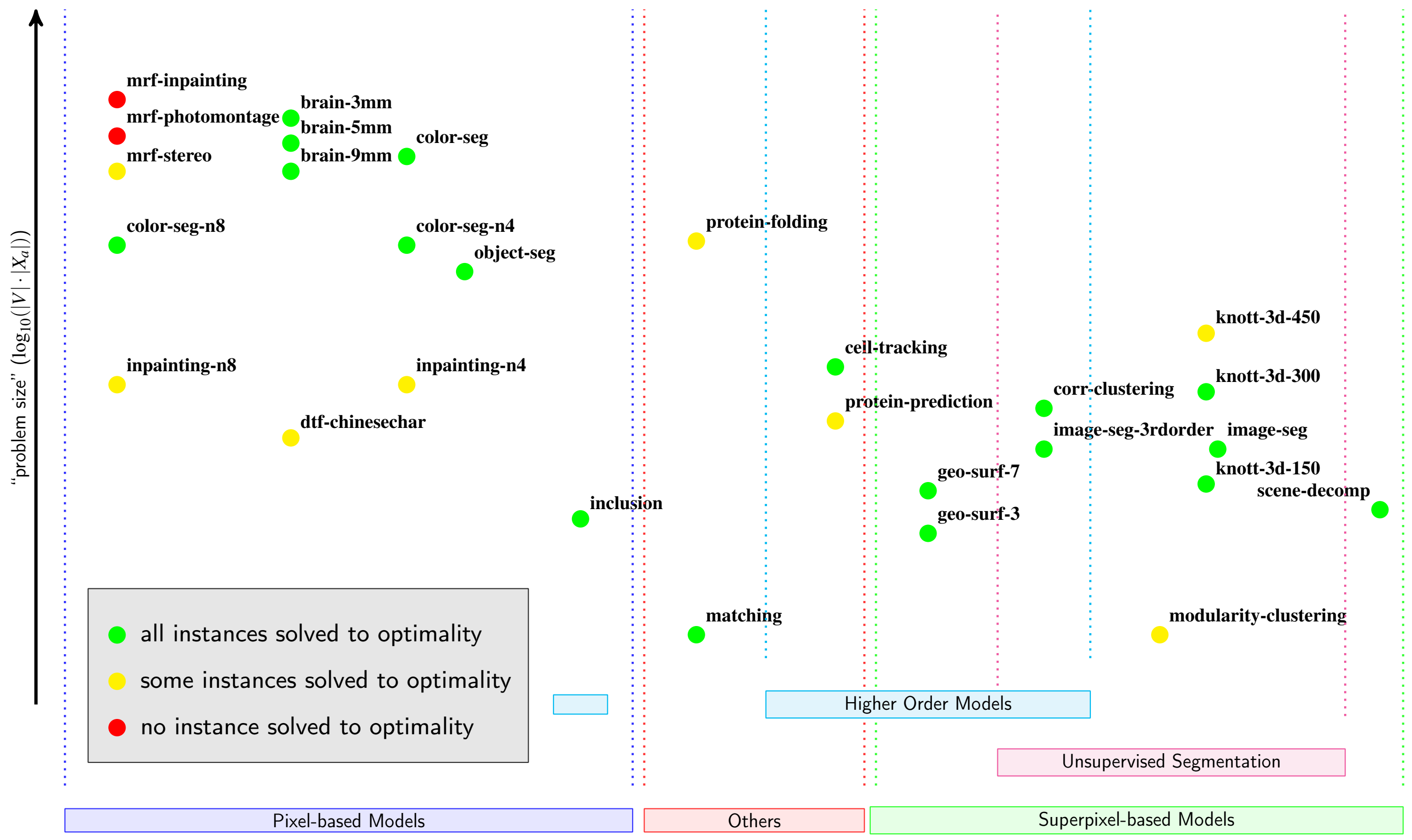

Figure 7: List of models used in the benchmark. The x-axis groups the model into specific classes and the y-axis reflects the size of the models. Datasets for which we can calculate the optimal solution within 1 hour for each instances are marked green, for which some solved to optimality within 1 hour in yellow, and those which are unsolved within 1 hour so far red. If we gave combinatorial methods more time for optimization, say 10 hours, we would be able to solve some more models to optimality [61]. We were able to find better solutions than reported in [66] for the models there considered, even we were not able to solve a single instance in mrf-photomontage or mrf-inpainting within 1 hour. Non surprisingly, larger models are usually harder than smaller ones. If special properties can be used, as for Potts models, solvers scale better. Also small models can be hard if large interactions exist, as for dtf-chinesechar. While our study gives some indications how the "hardness" of an instance could be estimated, a principle measure is so far not known.

For some models, we are able to evaluate the output also with respect to an application specific measurement, cf. Tab. 1. This addresses the question whether the absolute differences between algorithms are relevant for a particular application.

## 6.1 Pixel/Voxel-Based Models

**Stereo Matching** (mrf-stereo). We now consider three instances of a graphical model for the stereo matching problem in vision. Results are shown in Tab. 2. It can be seen from these results that two instances were solved to optimality. Only for the instance *teddy* in which variables have 60 labels, no integer solution could be verified as optimal. For the two instances for which optimal solutions were obtained, suboptimal approximations that were obtained significantly faster are not significant worse in terms of the two pixel accuracy (PA2), i.e. the number of pixels whose disparity error is less than or equal to two. On average, the solution obtained by BPS is 2% better in terms of the two-pixel accuracy (PA2) than solutions with smaller objective value.

**Inpainting** (mrf-inpainting). We now consider two instances of a graphical model for image inpainting. In these instances, every variable can attain 256 labels. Thus, efficient implementation is essential and only some implementations of approximative algorithms could be applied. Results are shown in Tab. 3 and Fig. 8. It can be seen from these results that TRWS outperforms move making methods. The best result is obtained by taking the solution provided by TRWS as the starting point for a local search by lazy flipping. While FastPD and $\alpha$-expansion converge faster than TRWS, their solution is significantly worse in terms of the objective value and also in terms of the mean color error (CE).

[5] As we deal with floating-point numbers and terminate algorithms if the gap between the current best integer solution and the lower bound is less than $10^{-5}$, we need to take the precision of floating-point operations into account when deciding whether an algorithm performed best or verified optimality. An output is taken to be the best and verified optimal if the difference is less than $10^{-5}$ in terms of its absolute value or less than $10^{-8}$ in terms of its relative value, see Fig. 13 for a formal definition.

| algorithm | runtime | value | bound | best | opt | PA2 |
|---|---|---|---|---|---|---|
| FastPD | 3.34 sec | 1614255.00 | $-\infty$ | 0 | 0 | 0.6828 |
| mrf-$\alpha$-Exp-TL | 10.92 sec | 1616845.00 | $-\infty$ | 0 | 0 | 0.6823 |
| mrf-$\alpha\beta$-Swap-TL | 10.14 sec | 1631675.00 | $-\infty$ | 0 | 0 | 0.6832 |
| ogm-LF-2 | 366.02 sec | 7396373.00 | $-\infty$ | 0 | 0 | 0.3491 |
| ogm-TRWS-LF2 | 439.75 sec | **1587043.67** | **1584746.53** | 0 | 0 | 0.6803 |
| mrf-LBP-TL | 287.62 sec | 1633343.00 | $-\infty$ | 0 | 0 | 0.6804 |
| mrf-BPS-TL | 238.70 sec | 1738696.00 | $-\infty$ | 0 | 0 | **0.7051** |
| ogm-LBP-0.95 | 3346.05 sec | 1664620.33 | $-\infty$ | 0 | 0 | 0.6720 |
| ogm-TRBP-0.95 | 3605.25 sec | 1669347.00 | $-\infty$ | 0 | 0 | 0.6710 |
| mrf-TRWS-TAB | 1432.57 sec | 1587681.33 | 1584725.98 | 0 | 0 | 0.6806 |
| mrf-TRWS-TL | 227.67 sec | 1587928.67 | **1584746.53** | 0 | 0 | 0.6803 |
| ogm-ADSAL | 3600.54 sec | 1590304.67 | 1584501.22 | 1 | 1 | 0.6803 |
| ogm-BUNDLE-A+ | 2180.49 sec | 1648854.67 | 1584056.35 | 1 | 1 | 0.6803 |
| ogm-CombiLP | 969.33 sec | 1587560.67 | 1584724.04 | **2** | **2** | 0.6809 |

Table 2: *mrf-stereo* (3 instances): On average, TRWS-LF2 and CombiLP afford the best solutions. FastPD is the fastest algorithm. Solutions obtained by BPS are better in terms of the two-pixel accuracy (PA2) than solutions with lower objective value. Storing the functions of a graphical model explicitly, as value tables, instead of as implicit functions, slows algorithms down, as can be seen for TRWS.

| algorithm | runtime | value | bound | best | opt | CE |
|---|---|---|---|---|---|---|
| FastPD | 8.47 sec | 32939430.00 | $-\infty$ | 0 | 0 | 14.7 |
| mrf-$\alpha$-Exp-TL | 54.21 sec | 27346899.00 | $-\infty$ | 0 | 0 | 11.3 |
| mrf-$\alpha\beta$-Swap-TL | 111.13 sec | 27154283.50 | $-\infty$ | 0 | 0 | 12.0 |
| ogm-TRWS-LF1 | 736.49 sec | 26464015.00 | 26462450.59 | 0 | 0 | 10.9 |
| ogm-TRWS-LF2 | 3009.52 sec | **26463829.00** | **26462450.59** | **1** | 0 | 10.9 |
| mrf-LBP-TL | 666.19 sec | 26597364.50 | $-\infty$ | 0 | 0 | **10.5** |
| mrf-BPS-TL | 644.15 sec | 26612532.50 | $-\infty$ | 0 | 0 | 12.0 |
| mrf-TRWS-TL | 614.05 sec | 26464865.00 | **26462450.59** | 0 | 0 | 10.9 |
| ogm-ADSAL | 3765.37 sec | 26487395.50 | 26445260.33 | 0 | 0 | 10.9 |
| ogm-CombiLP | 49672.26 sec | 26467926.00 | 26461874.39 | **1** | 0 | 10.9 |

Table 3: *mrf-inpainting* (2 instances): The best results are obtained by TRWS-LF2. $\alpha$-Expansion is a faster but worse alternative. The smallest color error (CE) was obtained by BPS.

**Photomontage** (mrf-photomontage). We now consider two instances of graphical models for photomontage. Results are shown in Tab. 4. It can be seen from these results that move making algorithms outperform algorithms based on linear programming relaxations. This observation is explained by the fact that the second-order factors are more discriminative in this problem than the first-order factors. Therefore, the LP relaxation is loose and thus, finding good primal solutions (rounding) is hard.

**Color Segmentation** (col-seg-n4/-n8). We now consider nine instances of a graphical model for color segmentation. These are Potts models with 76.800 variables and few labels and a 4-connected or 8-connected grid graph. Results are shown in Tab. 5 and Fig. 9. It can be seen form these results that all instances could be solved by the multicut algorithm and CombiLP, both of which verified optimality of the respective solutions. LP relaxations over the local polytope are tight for 7 instances and are overall better than other approximations. FastPD, $\alpha$-expansion and $\alpha\beta$-swap converged to somewhat worse but still reasonable solutions very quickly. For the hardest instance, algorithms based on multiway cuts did not find a solution within one hour. It can be seen from Fig. 9 that approximate solutions differ, especially at the yellow feathers around the neck.

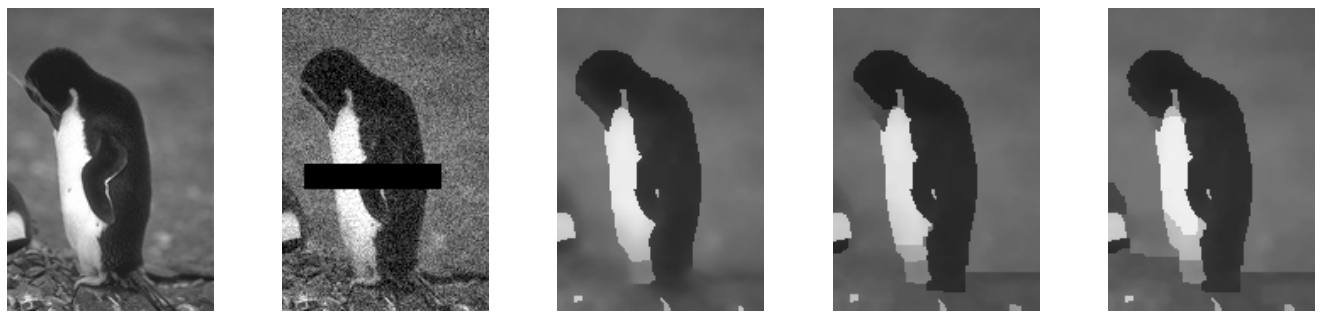

(a) image (b) input (c) TRWS (d) FastPD (e) $\alpha$-Exp

Figure 8: *mrf-inpainting*: Depicted above is one example of image inpainting. From the original image (a), a box is removed and noise is added to the remaining part (b). The result of inpainting and denoising by means of TRWS (c) is better than that of FastPD (d) and $\alpha$-expansion (e) which show artifacts.

| algorithm | runtime | value | bound | best | opt |
|---|---|---|---|---|---|
| mrf-$\alpha$-Exp-TAB | **7.54** sec | **168284.00** | $-\infty$ | **2** | 0 |
| mrf-$\alpha\beta$-Swap-TAB | 8.42 sec | 200445.50 | $-\infty$ | 0 | 0 |
| ogm-TRWS-LF1 | 300.21 sec | 1239959.00 | **166827.12** | 0 | 0 |
| ogm-TRWS-LF2 | 390.34 sec | 735193.00 | **166827.12** | 0 | 0 |
| mrf-LBP-TAB | 686.61 sec | 438611.00 | $-\infty$ | 0 | 0 |
| mrf-BPS-TAB | 167.49 sec | 2217579.50 | $-\infty$ | 0 | 0 |
| mrf-TRWS-TAB | 172.20 sec | 1243144.00 | **166827.07** | 0 | 0 |
| ogm-ADSAL | 3618.86 sec | 182045.50 | 165979.33 | 0 | 0 |

Table 4: *mrf-photomontage* (2 instances): For these instances, $\alpha$-Expansion is clearly the first choice. Due to the lack of unary data-terms, the LP relaxation is weak and rounding is hard.

**Color Segmentation** (color-seg). We now consider three instances of a graphical model for color segmentation provided by Alahari et al. [3]. Results are shown in Tab. 6. It can be seen from these results that the local polytope relaxation is tight. Approximate algorithms find the optimal solution for two instances and a near optimal solution for one instance of the problem. When Kovtun's method is used to reduce the problem size–which works well for this problem–the reduced problem can be solved easily and overall faster than with approximative algorithms alone.

**Object Segmentation** (object-seg). We now consider five instances of a graphical model for object segmentation provided by Alahari et al. [3]. Results are shown in Tab. 7. As for the color segmentation instances, the local polytope relaxation is tight. Compared to TRWS, FastPD is 10 times faster and worse in terms of the objective value only for 1 instance and only marginally. Furthermore, the pixel accuracy (PA) for results of FastPD and $\alpha$-expansion is slightly better than for optimal solutions. For instances like these which are large and easy to solve, combinatorial algorithms offer no advantages in practice.

**Inpainting** (inpainting-n4/n8). We now consider four synthetic instances of a graphical model for image inpainting, two instances with a 4-connected grid graph and two instances with an 8-connected grid graph. For each graph structure, one instance has strong first-order factors and the other instance (with postfix 'inverse') is constructed such that, for every variable, a first-order factors assigns to same objective value to two distinct labels. Results are shown in Tab. 8 and Fig. 10. It can be seen from these results that even Potts models can give rise to hard optimization problems, in particular if the first-order factors do not discriminate well between labels. Moreover, it can be seen from Fig. 10 that increasing the neighborhood-system helps to avoid discretization artefacts. However, even with an 8-neighborhood, the model favors 135° over 120° angles, as can be seen in Fig. 10(c).

| algorithm | runtime | value | bound | best | opt |
|---|---|---|---|---|---|
| FastPD | **0.27** sec | 20034.80 | $-\infty$ | 0 | 0 |
| mrf-$\alpha$-Exp-TL | 0.94 sec | 20032.61 | $-\infty$ | 0 | 0 |
| mrf-$\alpha\beta$-Swap-TL | 0.68 sec | 20059.02 | $-\infty$ | 0 | 0 |
| ogm-FastPD-LF2 | 9.64 sec | 20033.21 | $-\infty$ | 0 | 0 |
| ogm-ICM | 2.09 sec | 26329.45 | $-\infty$ | 0 | 0 |
| ogm-TRWS-LF2 | 9.90 sec | 20012.17 | 20012.14 | 7 | 7 |
| mrf-LBP-TL | 46.45 sec | 20053.25 | $-\infty$ | 0 | 0 |
| mrf-BPS-TL | 24.75 sec | 20094.03 | $-\infty$ | 0 | 0 |
| ogm-LBP-0.5 | 1573.21 sec | 20054.27 | $-\infty$ | 0 | 0 |
| MCR-TC-MTC | 464.79 sec | 20424.79 | 19817.76 | 6 | 6 |
| mrf-TRWS-TL | **8.47** sec | 20012.18 | **20012.14** | 8 | 7 |
| ogm-ADSAL | 983.73 sec | 20012.15 | 20012.14 | 8 | 7 |
| TRWS-pct | 135.38 sec | 20012.17 | 20012.14 | 7 | 7 |
| MCI-TC-MTC-TCI | 564.17 sec | 20416.16 | 19817.76 | 8 | 8 |
| MCI-pct | 507.98 sec | 20889.89 | $-\infty$ | 8 | 8 |
| ogm-CombiLP | 42.11 sec | **20012.14** | **20012.14** | **9** | **9** |

Table 5: *color-seg-n4* (9 instances): TRWS gives optimal or nearly optimal results on all instances. CombiLP also solves all problems. MCI solves all but one instance.

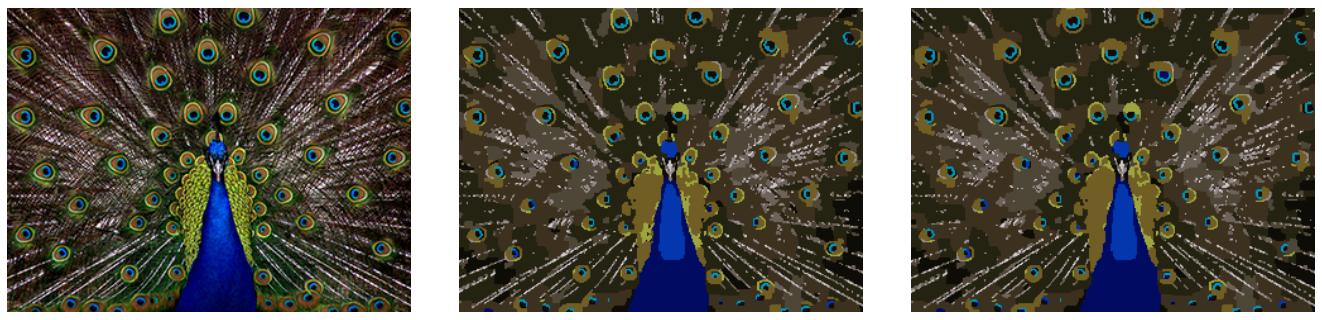

(a) input data (b) CombiLP (c) FastPD

Figure 9: *color-seg-n4*: For the hardest instance of the color segmentations problems, the differences between the optimal solution (b) and the approximate ones, here exemplary for FastPD (c), are small but noticeable.

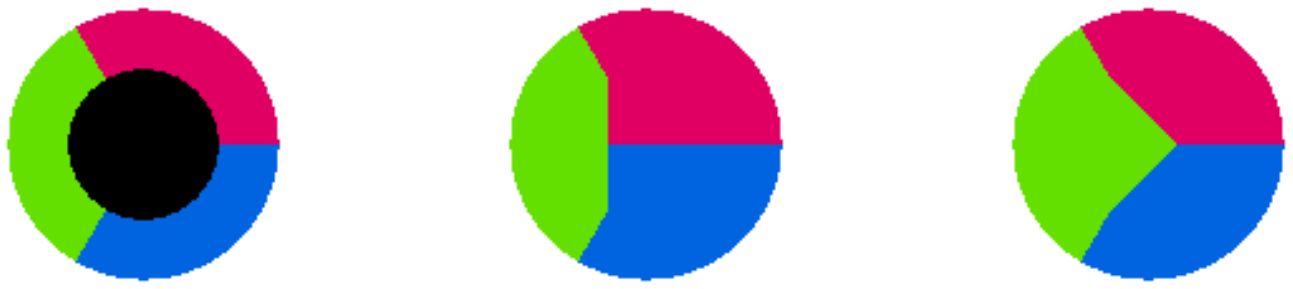

(a) input data (b) result for N4 (c) result for N8

Figure 10: *inpainting*. Depicted above are solutions (b) and (c) of synthetic instances of the inpainting problem (a). It can be seen that discretization artifacts due to the 4-neighborhood (b) are smaller than discretization artifacts due to the 8-neighborhood (c).

| algorithm | runtime | value | bound | best | opt |
|---|---|---|---|---|---|
| $\alpha$-Exp-pct | **1.07** sec | 308472274.33 | $-\infty$ | **3** | 0 |
| $\alpha$-Exp-VIEW | 6.12 sec | 308472275.67 | $-\infty$ | 2 | 0 |
| FastPD | **0.31** sec | 308472275.00 | $-\infty$ | 2 | 0 |
| ogm-LF-2 | 13.70 sec | 309850181.00 | $-\infty$ | 0 | 0 |
| $\alpha\beta$-Swap-VIEW | 6.54 sec | 308472292.33 | $-\infty$ | 2 | 0 |
| BPS-TL | 70.81 sec | 308733349.67 | $-\infty$ | 0 | 0 |
| ogm-LBP-0.5 | 983.26 sec | 308492950.67 | $-\infty$ | 0 | 0 |
| ogm-LBP-0.95 | 328.49 sec | 308494213.33 | $-\infty$ | 0 | 0 |
| MCR-TC-MTC | 84.99 sec | **308472274.33** | **308472274.33** | **3** | **3** |
| ogm-ADSAL | 2505.06 sec | **308472274.33** | 308472274.31 | **3** | 2 |
| TRWS-TL | 94.15 sec | 308472310.67 | 308472270.43 | 2 | 1 |
| TRWS-pct | **1.08** sec | 308472290.67 | **308472274.33** | 2 | 2 |
| MCI-TC-TCI | 92.90 sec | **308472274.33** | **308472274.33** | **3** | **3** |
| MCI-pct | **1.26** sec | **308472274.33** | **308472274.33** | **3** | **3** |
| ogm-CombiLP | 433.98 sec | **308472274.33** | **308472274.33** | **3** | **3** |

Table 6: *color-seg* (3 instances): For all instances, the local polytope relaxation is tight. Nevertheless, the fixed point of TRWS is suboptimal and for one instance, ADSAL does not converge within 1 hour. MCI-pct provides verified solutions as fast as $\alpha$-Expansion and FastPD which do not provide optimality certificates.

| algorithm | runtime | value | bound | best | opt | PA |
|---|---|---|---|---|---|---|
| FastPD | **0.13** sec | 31317.60 | $-\infty$ | 4 | 0 | **0.8856** |
| mrf-$\alpha$-Exp-TL | 0.27 sec | 31317.60 | $-\infty$ | 4 | 0 | **0.8856** |
| mrf-$\alpha\beta$-Swap-TL | 0.27 sec | 31318.70 | $-\infty$ | 3 | 0 | 0.8835 |
| mrf-LBP-TL | 30.06 sec | 32400.01 | $-\infty$ | 0 | 0 | 0.8439 |
| mrf-BPS-TL | 11.37 sec | 35775.27 | $-\infty$ | 0 | 0 | 0.7992 |
| MCR-TC-MTC | 482.85 sec | 32034.47 | 31317.23 | 4 | 4 | 0.8842 |
| mrf-TRWS-TL | **2.25** sec | **31317.23** | **31317.23** | **5** | **5** | 0.8800 |
| ogm-ADSAL | 115.52 sec | **31317.23** | **31317.23** | **5** | **5** | 0.8800 |
| TRWS-pct | **1.15** sec | **31317.23** | **31317.23** | **5** | **5** | 0.8800 |
| MCI-TC-MTC-TCI | 580.41 sec | **31317.23** | **31317.23** | **5** | **5** | 0.8800 |
| MCI-TC-TCI | 492.34 sec | **31317.23** | **31317.23** | **5** | **5** | 0.8800 |
| MCI-pct | 73.56 sec | **31317.23** | **31317.23** | **5** | **5** | 0.8800 |
| ogm-CombiLP | 37.47 sec | **31317.23** | **31317.23** | **5** | **5** | 0.8800 |

Table 7: *object-seg* (5 instances): For instances like these which are large and for which the local polytope relaxation is tight, combinatorial algorithms offer no advantages in practice.

**Brain Segmentation** (brain-3/5/7mm). We now consider twelve instances of a graphical model for segmenting MRI scans of human brains defined by four simulated scans at three different resolutions. These instances have $10^5 - 10^6$ variables. For such large instances, the efficient data structures are helpful. Results are shown in Tab. 9. It can be seen from these results that TRWS provides tight lower bounds but suboptimal approximate solutions which shows that the rounding problem remains hard. Multiway cut cannot be applied directly because the instances are too large. However, with model reduction as pre-processing, MCI-ptc is the only algorithm that could solve all instances within one hour. FastPD terminates in 1/10 of the runtime, providing approximations which are worse in terms of the objective value but reasonable in the application domain.

**DTF Chinese Characters Inpainting** (dtf-chinesechar). We now consider 100 instances of a graphical model for Chinese character inpaining. In contrast to all models considered so far, these instances are decision tree fields (DTF) which are learned in a discriminative fashion. This gives rise to frustrated cycles which render the inference problem hard. Results are shown in Fig. 11 and Tab. 10. It can be seen from these results that the local polytope relaxation is loose. Moreover, it can be seen that instead of applying a combinatorial algorithm directly, it is beneficial to first reduce the problem. Here, MCBC-pct performs best, verifying optimality of 56 of 100 instances. In shorter time, sequential belief propagation (BPS) and lazy flipping give good results. With respect to the pixel accuracy (PA) in the inpainting-region, BPS is with 67,15% correct in-painted pixels, better than MCBC-pct which has PA of 66,24%.

| algorithm | runtime | value | bound | best | opt |
|---|---|---|---|---|---|
| FastPD | 0.03 sec | 454.75 | $-\infty$ | 1 | 0 |
| mrf-$\alpha$-Exp-TL | **0.02** sec | **454.35** | $-\infty$ | **2** | 0 |
| mrf-$\alpha\beta$-Swap-TL | 0.02 sec | 454.75 | $-\infty$ | 1 | 0 |
| ogm-TRWS-LF2 | 1.98 sec | 489.30 | 448.09 | 1 | **1** |
| mrf-LBP-TL | 4.68 sec | 475.56 | $-\infty$ | 1 | 0 |
| mrf-BPS-TL | 1.63 sec | **454.35** | $-\infty$ | **2** | 0 |
| mrf-TRWS-TL | 0.96 sec | 490.48 | 448.09 | 1 | **1** |
| ogm-ADSAL | 101.85 sec | 454.75 | **448.27** | 1 | **1** |
| ogm-SG-A | 36.63 sec | 455.25 | 447.76 | 1 | **1** |
| ogm-SG-A+ | 29.85 sec | 455.25 | 441.36 | 1 | **1** |
| ogm-SG-A- | 57.99 sec | **454.35** | 447.89 | **2** | 0 |
| TRWS-pct | 6.42 sec | 489.30 | 448.10 | 1 | **1** |
| MCI-TC-MTC-TCI | 1821.99 sec | 511.69 | 448.07 | 1 | **1** |
| ogm-CombiLP | 1867.58 sec | 461.81 | 446.66 | 1 | **1** |

Table 8: **inpainting-n4** (2 instances): One instance is easy because the LP relaxations is tight. The other instance is designed to be hard. On average, $\alpha$-Expansion and BPS perform best. Subgradient algorithms with small step-size give the best primal solutions, but dual convergence is slow. Combinatorial algorithms cannot solve the hard instance within 1 hour.

| algorithm | runtime | value | bound | best | opt |
|---|---|---|---|---|---|
| $\alpha$-Exp-pct | 10.16 sec | 19088999.75 | $-\infty$ | 0 | 0 |
| $\alpha$-Exp-VIEW | 103.27 sec | 19089080.00 | $-\infty$ | 0 | 0 |
| FastPD | **2.22** sec | 19089484.75 | $-\infty$ | 0 | 0 |
| ogm-TRWS-LF2 | 220.80 sec | 19087628.00 | **19087612.50** | 0 | 0 |
| BPS-TL | 453.81 sec | 19090723.25 | $-\infty$ | 0 | 0 |
| ogm-ADSAL | 3612.32 sec | 19087690.00 | 19087612.49 | 0 | 0 |
| TRWS-TL | 152.94 sec | 19087730.25 | **19087612.50** | 0 | 0 |
| TRWS-pct | 23.21 sec | 19087728.50 | **19087612.50** | 0 | 0 |
| MCI-pct | **29.20** sec | **19087612.50** | **19087612.50** | **4** | **4** |
| ogm-CombiLP | 2761.11 sec | 19087646.75 | **19087612.50** | 2 | 2 |

Table 9: *brain-5mm* (4 instances): Although the local polytope relaxation is tight for these instances, rounding is not trivial. Only MCI-pct is able to solve all instances within 1 hour. FastPD is 10 times faster but solutions are worse in terms of the objective value. For real world applications, these solutions might, however, be sufficient.

| algorithm | runtime | value | bound | best | opt | PA |
|---|---|---|---|---|---|---|
| ogm-LF-1 | **0.44** sec | $-49516.08$ | $-\infty$ | 1 | 0 | 0.5725 |
| ogm-LF-2 | 15.28 sec | $-49531.11$ | $-\infty$ | 7 | 0 | 0.6003 |
| ogm-LF-3 | 963.78 sec | $-49535.37$ | $-\infty$ | 17 | 0 | 0.6119 |
| BPS-TAB | 78.65 sec | $-49537.08$ | $-\infty$ | 30 | 0 | **0.6715** |
| ogm-ADSAL | 1752.19 sec | $-49524.32$ | $-50119.39$ | 1 | 0 | 0.6487 |
| QPBO | 0.20 sec | $-49501.95$ | $-50119.38$ | 0 | 0 | 0.5520 |
| TRWS-pct | 7.58 sec | $-49496.76$ | $-50119.38$ | 2 | 0 | 0.5636 |
| ogm-ILP-pct | 3556.36 sec | $-49547.48$ | $-50062.14$ | 62 | 0 | 0.6540 |
| MCBC-pct | 2053.89 sec | **$-49550.10$** | **$-49612.38$** | **83** | **56** | 0.6624 |
| ogm-ILP | 3569.92 sec | $-49533.85$ | $-50096.11$ | 8 | 0 | 0.6399 |
| SA | – | $-49533.02$ | $-\infty$ | 13 | 0 | 0.6541 |

Table 10: *dtf-chinesechar* (100 instances): Best results are obtained by combinatorial methods after model reduction. MCBC use special cutting plane methods which lead to tighter relaxations and better lower bounds. When a shorter running time is needed the Lazy Flipper and BPS are alternatives.

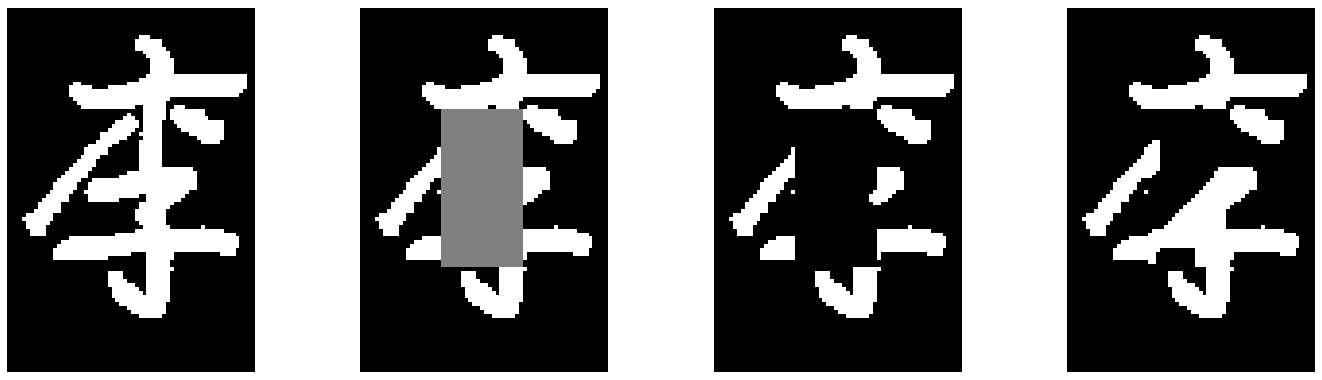

(a) GT (b) data (c) TRWS (d) MCBC

Figure 11: *dtf-chinesechar*: Depicted above is one example of the Chinese character inpainting problem. The purpose of the model is to reconstruct the original image (a), more precisely, the mask (b), from the rest of the image. It can be seen in (c) that TRWS labels the complete inpainting area as background. MCBC finds the optimal solution which reflects the full potential of decision tree fields for this application.

**Color Segmentation with Inclusion** (inclusion). We now consider ten instances of a higher-order graphical models for color segmentation with inclusion. Due to the higher-order factors, some methods, e.g. TRWS, are no longer applicable. Results are shown in Tab. 11. It can be seen from these results that the local polytope relaxation is quite tight. However, standard rounding procedures do not yield good integer solutions. Overall, the ILP solver performs best. For larger problems, ILPs might not scale well. In such a case, LBP followed by lazy flipping is a good alternative. The best pixel accuracy (PA) is obtained by Bundle-methods, which is 0.1% better than optimal results that have a pixel accuracy of 94.96%.

## 6.2 Superpixel-Based Models

Graphical models defined with respect to an adjacency graph of superpixels have fewer variables than graphical models defined with respect to the pixel grid graph. The relative difference in the number of variables can be several orders of magnitude. Thus, combinatorial algorithms are more attractive for superpixel-based models.

**Scene decomposition** (scene-decomposition). We now consider 715 instances of a graphical model for scene-decomposition. These instances have between 150 and 208 variables. Results are shown in Tab. 12. It can be seen from these results that the differences between the solutions of the different methods are marginal, both in terms of objective value and in terms of pixel accuracy (PA), i.e. the percentage of correctly labeled pixels. While TRWS is the fastest method, CombiLP is the fastest method that solves all problems to optimality. The best results in terms of pixel accuracy are given by TRBP.

| algorithm | runtime | value | bound | best | opt | PA |
|---|---|---|---|---|---|---|
| ogm-LBP-LF2 | 95.66 sec | 1400.61 | $-\infty$ | 7 | 0 | 0.9495 |
| ogm-LF-1 | 0.01 sec | 1556.20 | $-\infty$ | 0 | 0 | 0.6206 |
| ogm-LF-2 | 0.14 sec | 1476.39 | $-\infty$ | 0 | 0 | 0.7630 |
| ogm-LF-3 | 1.69 sec | 1461.23 | $-\infty$ | 0 | 0 | 0.8011 |
| ogm-LBP-0.5 | 96.11 sec | 2100.61 | $-\infty$ | 7 | 0 | 0.9487 |
| ogm-LBP-0.95 | 104.38 sec | 2700.74 | $-\infty$ | 8 | 0 | 0.9492 |
| ogm-TRBP-0.5 | 103.46 sec | 1900.84 | $-\infty$ | 5 | 0 | 0.9491 |
| ogm-TRBP-0.95 | 100.35 sec | 2600.73 | $-\infty$ | 8 | 0 | 0.9481 |
| ADDD | 7.24 sec | 3400.81 | 1400.31 | 1 | 1 | 0.9479 |
| MPLP | 7.47 sec | 4000.44 | 1400.30 | 2 | 1 | 0.9479 |
| ogm-BUNDLE-H | 137.47 sec | 1400.76 | 1400.32 | 3 | 1 | 0.9496 |
| ogm-BUNDLE-A- | 151.85 sec | 1400.68 | 1400.30 | 4 | 0 | **0.9506** |
| ogm-SG-A- | 154.05 sec | 26797.36 | 1343.02 | 0 | 0 | 0.8518 |
| ogm-LP-LP | 20.98 sec | 3900.59 | 1400.33 | 1 | 1 | 0.9484 |
| ogm-ILP | **6.54** sec | **1400.57** | **1400.57** | **10** | **10** | **0.9496** |

Table 11: *inclusion* (10 instances): Only the commercial ILP software solves all instances of this dataset. While the local polytope relaxation is quite tight, rounding is still hard. Lazy flipping can help to correct some rounding errors.

The PA of TRBP is with 77.08% slightly better that the PA for optimal solutions.

| algorithm | runtime | value | bound | best | opt | PA |
|---|---|---|---|---|---|---|
| ogm-LBP-LF2 | 0.31 sec | −866.76 | $-\infty$ | 576 | 0 | 0.7699 |
| ogm-LF-3 | 0.99 sec | −866.27 | $-\infty$ | 420 | 0 | 0.7699 |
| ogm-TRWS-LF2 | **0.01** sec | −866.93 | −866.93 | 714 | 712 | 0.7693 |
| BPS-TAB | 0.12 sec | −866.73 | $-\infty$ | 566 | 0 | 0.7701 |
| ogm-LBP-0.95 | 0.09 sec | −866.76 | $-\infty$ | 580 | 0 | 0.7696 |
| ogm-TRBP-0.95 | 0.61 sec | −866.84 | $-\infty$ | 644 | 0 | **0.7708** |
| ADDD | 0.09 sec | −866.92 | −866.93 | 701 | 697 | 0.7693 |
| MPLP | 0.06 sec | −866.91 | −866.93 | 700 | 561 | 0.7693 |
| MPLP-C | 0.07 sec | −866.92 | −866.93 | 710 | 567 | 0.7693 |
| ogm-ADSAL | 0.17 sec | −866.93 | −866.93 | 714 | 712 | 0.7693 |
| ogm-BUNDLE-H | 0.76 sec | **−866.93** | −866.93 | **715** | 673 | 0.7693 |
| ogm-BUNDLE-A+ | 0.21 sec | **−866.93** | −866.93 | **715** | 712 | 0.7693 |
| ogm-SG-A+ | 0.21 sec | −866.92 | −866.93 | 711 | 707 | 0.7694 |
| ogm-LP-LP | 0.28 sec | −866.92 | −866.93 | 712 | 712 | 0.7693 |
| TRWS-TAB | **0.01** sec | −866.93 | −866.93 | 714 | 712 | 0.7693 |
| ADDD-BB | 0.15 sec | **−866.93** | **−866.93** | **715** | **715** | 0.7693 |
| ogm-CombiLP | **0.05** sec | **−866.93** | **−866.93** | **715** | **715** | 0.7693 |
| ogm-ILP | 0.24 sec | **−866.93** | **−866.93** | **715** | **715** | 0.7693 |
| BRAOBB-1 | 19.07 sec | −866.90 | $-\infty$ | 670 | 0 | 0.7688 |

Table 12: *scene-decomposition* (715 instances): Almost all algorithms provide optimal or nearly optimal solutions. Also in terms of pixel accuracy (PA), the difference between these solutions is marginal. The PA of optimal solutions is 0.2% worse than the PA of the suboptimal solutions found by TRBP. TRWS is the fastest algorithm, followed by CombiLP.

**Geometric Surface Labeling** (geo-surf-3/7). We now consider $2 \times 300$ instances of higher-order graphical models for geometric surface labeling. Results are shown in Tab. 13. It can be seen from these results that the local polytope relaxation is very tight and the commercial ILP solver performs best. With increasing number of labels (from three to seven), non-commercial combinatorial algorithms suffer more than the commercial solver which performs well across the entire range, finding optimal solutions faster than approximative algorithms take to converge. For *geo-surf-7*, only $\alpha$-Fusion is significant faster, but the approximate solutions are also worse. In terms of pixel accuracy, suboptimal approximations are better than optimal solutions. LBP-LF2 has a 0.07% and 4.59% higher pixel accuracy (PA) than the optimal labeling for *geo-surf-3* and *geo-surf-7*, respectively. This indicates that, at least for *geo-surf-7*, the model does not reflect the pixel accuracy loss and can potentially be improved.

| algorithm | runtime | value | bound | best | opt | PA |
|---|---|---|---|---|---|---|
| $\alpha$-Fusion | **0.02** sec | 477.83 | $-\infty$ | 257 | 0 | 0.6474 |
| ogm-LBP-LF2 | 3.97 sec | 498.44 | $-\infty$ | 66 | 0 | 0.6988 |
| ogm-LBP-0.5 | 3.07 sec | 498.45 | $-\infty$ | 67 | 0 | **0.6988** |
| ogm-TRBP-0.5 | 35.79 sec | 486.42 | $-\infty$ | 128 | 0 | 0.6768 |
| ADDD | 0.66 sec | 476.95 | 476.94 | 296 | 293 | 0.6531 |
| MPLP | 1.83 sec | 477.56 | 476.94 | 278 | 195 | 0.6529 |
| MPLP-C | 2.00 sec | 477.34 | 476.95 | 282 | 198 | 0.6529 |
| ogm-BUNDLE-H | 76.77 sec | 476.95 | 476.86 | 299 | 180 | 0.6529 |
| ogm-BUNDLE-A+ | 59.55 sec | 476.95 | 476.91 | 299 | 239 | 0.6529 |
| ogm-SG-A+ | 56.57 sec | 477.29 | 476.75 | 295 | 250 | 0.6533 |
| ogm-LP-LP | 3.07 sec | 476.95 | 476.94 | 299 | 299 | 0.6530 |
| ogm-ILP | **1.01** sec | **476.95** | **476.95** | **300** | **300** | 0.6529 |
| BRAOBB-1 | 981.94 sec | 479.81 | $-\infty$ | 214 | 0 | 0.6602 |

Table 13: *geo-surf-7* (300 instances): The commercial ILP solver performs best for these small models. The local polytope relaxation is tight for almost all instances. While $\alpha$-Fusion is very fast, the solutions it provides are significant worse than the optimal solution, for some instances, both in terms of the objective value and in terms of pixel accuracy (PA). The suboptimal solutions provided by LBP have a better PA than optimal solutions.

## 6.3 Partition Models

The properties of graphical models for unsupervised image segmentation, namely (1) absence of unary terms, (2) invariance to label-permutation, and (3) huge label spaces, exclude most common inference methods for graphical models. Most important, the invariance to label-permutation causes that the widely used local polytope relaxation is more or less useless. That why we will compare here solvers that are designed to make use of or can deal with this additional properties.

**Modularity Clustering** (modularity-clustering). We now consider six instances of graphical models for finding a clustering of a network that maximize the modularity. Results are shown in Tab. 14. The Kerninghan-Lin algorithm is an established, fast and useful heuristic for clustering networks with respect to their modularity. It can be seen from the table that local search by means of ICM or LF-1 does not find better feasible solution than the initial labeling (a single cluster). While multicut methods work well for small instances, they do not scale so well because the graph is fully connected. It can also be seen that odd-wheel constraints tighten the standard LP-relaxation (with only cycle constrains) significantly. Combined LP/ILP cutting-plane methods (MCI-CCFDB-CCIFD) is the overall fastest exact method for networks of moderate size. However, for the largest instances, even MCI-CCFDB-CCIFD does not converge within one hour.

**Image Segmentation** (image-seg). We now consider 100 instances of graphical models for image segmentation. These models differ from models for network-analysis in that the former are sparse. Results are shown in Tab. 15. It can be seen from these results that standard LP-relaxations with only cycle constraints (MCR-CC and MCR-CCFDB) work well and significant better than KL, both in terms of objective value and variation of information (VI). Adding odd-wheel constraints (MCR-CCFDB-OWC) gives almost no improvement. Pure integer cutting plane methods (MCI-CCI and MCI-CCIFD) provide optimal solutions faster than LP-based methods and KL. Using only facet defining constraints reduces the number of added constraints and give better runtimes for MCR and MCI. Visually, the results are similar and vary only locally, cf. Fig. 12, indicating that fast and scalable approximative algorithms might be useful in practice.

| algorithm | runtime | value | bound | best | opt |
|---|---|---|---|---|---|
| ogm-ICM | 0.13 sec | 0.0000 | $-\infty$ | 0 | 0 |
| ogm-KL | **0.01** sec | **−0.4879** | $-\infty$ | 3 | 0 |
| ogm-LF-1 | 0.04 sec | 0.0000 | $-\infty$ | 0 | 0 |
| MCR-CC | 26.62 sec | −0.4543 | −0.5094 | 1 | 1 |
| MCR-CCFDB | 2.31 sec | −0.4543 | −0.5094 | 1 | 1 |
| MCR-CCFDB-OWC | 602.27 sec | −0.4665 | −0.4960 | 5 | 5 |
| MCI-CCFDB-CCIFD | 601.02 sec | −0.4400 | **−0.5019** | **5** | **5** |
| MCI-CCI | 1206.33 sec | −0.4363 | −0.5158 | 4 | 4 |
| MCI-CCIFD | 1203.71 sec | −0.4274 | −0.5174 | 4 | 4 |

Table 14: *modularity-clustering* (6 instances): The largest instance cannot be solved by any variant of MCI within one hour. KL is robust, fast and better on large instances, leading to a better mean objective value.

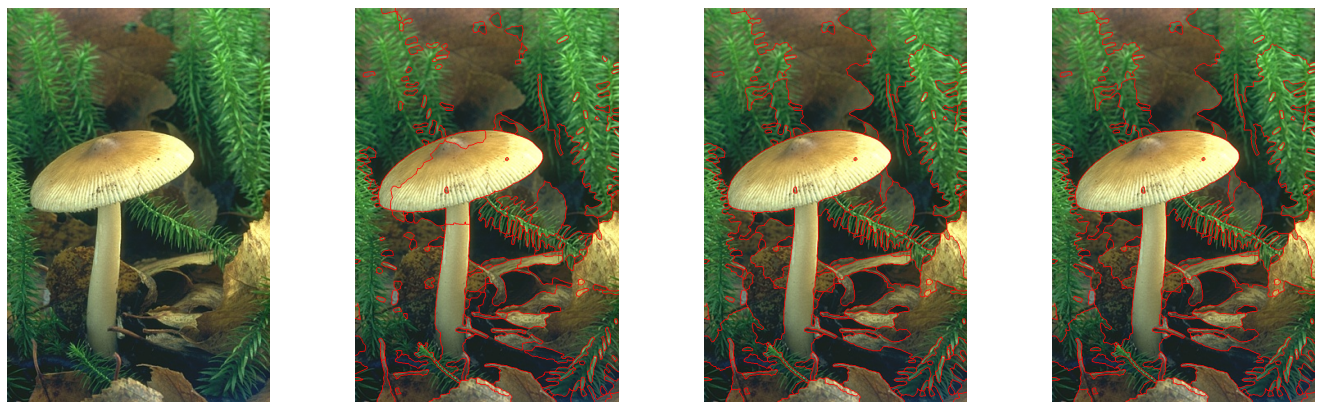

(a) image (b) KL (c) MC-CC (d) MC-CCI

Figure 12: *image-seg*: Depicted above is one example of the image segmentation problem. KL produces a segmentation which do not separate plants from background and constains incorrect boundaries. The ILP and LP-based multicut algorithms lead to reasonable results with differ only in the lower right part of the image.

| algorithm | runtime | value | bound | best | opt | VI |
|---|---|---|---|---|---|---|
| ogm-ICM | 4.24 sec | 4705.07 | $-\infty$ | 0 | 0 | 2.8580 |
| ogm-KL | **0.77** sec | 4608.55 | $-\infty$ | 0 | 0 | 2.6425 |
| ogm-LF-1 | 1.84 sec | 4705.01 | $-\infty$ | 0 | 0 | 2.8583 |
| MCR-CC | 2.40 sec | 4447.14 | 4442.34 | 36 | 35 | 2.5471 |
| MCR-CCFDB | 1.85 sec | 4447.14 | 4442.34 | 35 | 35 | 2.5469 |
| MCR-CCFDB-OWC | 2.20 sec | 4447.09 | 4442.34 | 35 | 35 | 2.5469 |
| MCI-CCFDB-CCIFD | 2.49 sec | **4442.64** | **4442.64** | **100** | **100** | **2.5367** |
| MCI-CCI | **2.33** sec | **4442.64** | **4442.64** | **100** | **100** | **2.5365** |
| MCI-CCIFD | 2.47 sec | **4442.64** | **4442.64** | **100** | **100** | **2.5367** |

Table 15: *image-seg* (100 instances): Variants of MCI solve all instances to optimality and are as fast as approximative algorithms.

**Higher-Order Image Segmentation** (image-seg-3rdorder). We now consider 100 instances of higher-order graphical models for image segmentation. Here, factors of order three are defined w.r.t. the angles between the tangents of contours at points in which contours meet. Results are shown in Tab. 16. It can be seen from these results that the standard LP relaxation (MCR-CCFDB) is no longer as tight as for the second-order models *image-seg*, cf. Tab. 15. Odd-wheel constraints (MCR-CCFDB-OWC) tighten this relaxation only marginally. Integer cutting plane methods (MCI) suffer from the weaker relaxations and need longer for optimization, but provide optimal solutions for all 100 instances. Consistent with the results reported in [6], we find that the VI of segmentations defined by optimal solutions of the third-order models is higher than the VI of segmentations defined by optimal solutions of the second-order models, indicating that either the hyper-parameter of the model needs to be estimated differently or the third-order terms are uninformative in this case.

| algorithm | runtime | value | bound | best | opt | VI |
|---|---|---|---|---|---|---|
| ogm-ICM | 7.12 sec | 6030.49 | $-\infty$ | 0 | 0 | 2.7089 |
| ogm-LF-1 | 2.71 sec | 6030.29 | $-\infty$ | 0 | 0 | 2.7095 |
| MCR-CC | 27.35 sec | 5822.42 | 5465.15 | 0 | 0 | 2.7723 |
| MCR-CCFDB | 16.24 sec | 5823.15 | 5465.15 | 0 | 0 | 2.7702 |
| MCR-CCFDB-OWC | 21.64 sec | 5823.64 | 5465.29 | 0 | 0 | 2.7702 |
| MCI-CCFDB-CCIFD | **43.75** sec | **5627.52** | **5627.52** | **100** | **100** | **2.6586** |
| MCI-CCI | 71.52 sec | 5628.31 | 5627.45 | 99 | 98 | 2.6588 |
| MCI-CCIFD | 59.88 sec | 5627.52 | 5627.52 | 99 | 100 | 2.6586 |

Table 16: *image-seg-3rdorder* (100 instances): Variants of MCI provide optimal solutions for all instances. Numerical problems arise only for some instances. Approximative methods are faster but results are worse.

**Hierarchical Image Segmentation** (hierarchical-image-seg): Next, we consider 715 instances of a graphical model for hierarchical image segmentation which include factors of orders up to 651. Results are shown in Tab. 17. For these instances, the standard LP relaxation (MCR-CC) is quite tight. Without odd-wheel constraints, 98 instances can be solved by the MCR. With odd-wheel constraints, 100 instances can be solved by the MCR, all with zero gap. For all 715 instances, the feasible solutions output by MCR are close to the optimum, both in terms of their objective as well as in terms of the VI. While MCR is 10 times faster than exact MCI algorithms, we emphasize that MCR was used to learn these instances which might be an advantage.

Due to well-known numerical issues, e.g. slackness introduced to improve numerical stability, bounds are not always tight for MCI. Parameters can be adjusted to overcome this problem for these particular instances, but such adjustments can render the solution of other models less numerically stable. Thus, we use the same parameters in all experiments.

**3D Neuron Segmentation** (knott-3d-150/300/450). We now consider $3 \times 8$ instances of graphical models for the segmentation of supervoxel adjacency graphs, i.e. for the segmentation of volume images. Results for instances on volume images of $300^3$ voxels are shown in Tab. 18. It can be seen from these results that MCR and MCI become slower. For the instances based on volume images of $300^3$ voxels, MCI-CCIFD solved instances within reasonable time. Without the restriction to facet defining constraints only, the system of inequalities grows too fast inside the cutting-plane procedure.

| algorithm | runtime | value | bound | best | opt | VI |
|---|---|---|---|---|---|---|
| ogm-ICM | 1.45 sec | −585.60 | −∞ | 0 | 0 | 2.6245 |
| ogm-LF-1 | 0.75 sec | −585.60 | −∞ | 0 | 0 | 2.6245 |
| MCR-CC | 0.08 sec | −626.76 | −628.89 | 162 | 98 | 2.0463 |
| MCR-CCFDB | 0.06 sec | −626.76 | −628.89 | 162 | 98 | 2.0463 |
| MCR-CCFDB-OWC | 0.06 sec | −626.77 | −628.89 | 164 | 100 | 2.0460 |
| MCI-CCFDB-CCIFD | **0.64** sec | **−628.16** | **−628.16** | **715** | **712** | **2.0406** |
| MCI-CCI | 1.28 sec | **−628.16** | −628.17 | **715** | 707 | **2.0406** |
| MCI-CCIFD | 1.26 sec | **−628.16** | −628.17 | **715** | 711 | **2.0406** |

Table 17: *hierarchical-image-seg* (715 instances): For these instances, the standard LP relaxation (MCR-CC) is quite tight. For all 715 instances, the feasible solutions output by MCR are close to the optimum, both in terms of their objective as well as in terms of the VI.

| algorithm | runtime | value | bound | best | opt | VI |
|---|---|---|---|---|---|---|
| ogm-ICM | 97.63 sec | −25196.51 | −∞ | 0 | 0 | 4.1365 |
| ogm-KL | 8.42 sec | −25553.79 | −∞ | 0 | 0 | 4.1380 |
| ogm-LF-1 | 34.11 sec | −25243.76 | −∞ | 0 | 0 | 4.1297 |
| MCR-CC | 1090.97 sec | −27289.63 | −27303.52 | 1 | 1 | 1.6369 |
| MCR-CCFDB | 309.11 sec | −27289.63 | −27303.52 | 1 | 1 | 1.6369 |
| MCR-CCFDB-OWC | 257.23 sec | −27301.42 | −27302.81 | 7 | 7 | 1.6373 |
| MCI-CCFDB-CCIFD | 251.28 sec | **−27302.78** | **−27302.78** | **8** | **8** | **1.6352** |
| MCI-CCI | 181.60 sec | **−27302.78** | −27305.02 | **8** | 7 | **1.6352** |
| MCI-CCIFD | **77.43** sec | **−27302.78** | **−27302.78** | **8** | **8** | **1.6352** |

Table 18: *knott-3d-300* (8 instances): MCI-CCIFD affords the best solutions. MCR-CCFDB-OWC affords good solutions as well but suffers form the fact that the separation procedure is more complex and time consuming for fractional solutions.

## 6.4 Other Models

**Matching** (matching). We now consider four instances of a graphical model for the matching problem. These instances have at most 21 variables. Results in shown in Tab. 19. It can be seen from these results that pure branch-and-bound algorithms, such as BRAOBB-1 or AStar, can solve these instances fast. In contrast, algorithm based on LP relaxations converge slower. The local polytope relaxation, used e.g. in MPLP, is loose because of the second-order soft-constraints used in this graphical model formulation of the matching problem. Thus, the rounding problem is hard. Adding additional cycle constraints, e.g. in MPLP-C, is sufficient to close the duality gap and obtain exact integer solutions. Another way to improve the objective value of poor integer solutions caused by violated soft-constraint is local search by lazy flipping.

**Cell Tracking** (cell-tracking). We now consider one instance of a graphical model for tracking biological cells in sequences of volume images. Results are shown in Tab. 20. It can be seen from these results that the ILP solver clearly outperforms all alternatives. Only the off-the-shelf LP-solver (LP-LP) manages to find a solution that satisfies all soft-constraints. Algorithms which solve the same relaxation, e.g.

| algorithm | runtime | value | bound | best | opt | MPE |
|---|---|---|---|---|---|---|
| ogm-LBP-LF2 | 0.34 sec | 38.07 | −∞ | 1 | 0 | 5.4469 |
| ogm-LF-2 | 0.46 sec | 40.79 | −∞ | 0 | 0 | 5.7689 |
| ogm-LF-3 | 20.29 sec | 39.81 | −∞ | 0 | 0 | 5.6346 |
| ogm-TRWS-LF2 | 0.47 sec | 33.31 | 15.22 | 0 | 0 | 3.1763 |
| BPS-TAB | 0.17 sec | 40.26 | −∞ | 0 | 0 | 4.9692 |
| ogm-LBP-0.5 | 0.01 sec | $105\cdot 10^9$ | −∞ | 0 | 0 | 6.0228 |
| ADDD | 2.31 sec | $105\cdot 10^9$ | 16.33 | 0 | 0 | 3.2429 |
| MPLP | 0.35 sec | $65\cdot 10^9$ | 15.16 | 0 | 0 | 3.1630 |
| MPLP-C | **4.63** sec | **21.22** | **21.22** | **4** | **4** | **0.0907** |
| ogm-ADSAL | 2371.04 sec | 34.58 | 16.04 | 0 | 0 | 2.8783 |
| ogm-LP-LP | 26.47 sec | $1025\cdot 10^9$ | 16.35 | 0 | 0 | 3.3482 |
| TRWS-TAB | 0.04 sec | 64.19 | 15.22 | 0 | 0 | 3.8159 |
| ogm-ASTAR | 7.20 sec | **21.22** | **21.22** | **4** | **4** | **0.0907** |
| ogm-CombiLP | 408.56 sec | **21.22** | **21.22** | **4** | **4** | **0.0907** |
| ogm-ILP | 1064.87 sec | **21.22** | **21.22** | **4** | **4** | **0.0907** |
| BRAOBB-1 | **2.25** sec | **21.22** | −∞ | **4** | 0 | **0.0907** |

Table 19: *matching* (4 instances): Pure branch & bound methods such as BRAOBB and AStar afford optimal solutions fast. Cycle constraints as used by MPLP-C tighten the relaxation sufficiently. For optimal solutions, the objective value correlates with the mean position error (MPE).

ADDD and MPLP, are slower and their rounded solutions violate soft-constraints. Lazy Flipping as a post-processing step can overcome this problem, as shown for LBP.

| algorithm | runtime | value | bound | best | opt |
|---|---|---|---|---|---|
| ogm-LBP-LF2 | 240.76 sec | 7515575.61 | −∞ | 0 | 0 |
| ogm-LF-3 | 2.38 sec | 8461693.24 | −∞ | 0 | 0 |
| ogm-LBP-0.95 | 236.29 sec | 307513873.84 | −∞ | 0 | 0 |
| ogm-TRBP-0.95 | 246.10 sec | 107517017.88 | −∞ | 0 | 0 |
| ADDD | 13.28 sec | 34108661574.79 | 6206883.23 | 0 | 0 |
| MPLP | 583.47 sec | 107514359.61 | 7513851.52 | 0 | 0 |
| ogm-LP-LP | **4.66** sec | 7516359.61 | 7513851.52 | 0 | 0 |
| ADDD-BB | 55077.04 sec | **7514421.21** | 7411393.72 | **1** | 0 |
| ogm-ILP-pct | **13.40** sec | **7514421.21** | **7514421.21** | **1** | **1** |
| ogm-ILP | **13.78** sec | **7514421.21** | **7514421.21** | **1** | **1** |

Table 20: *cell-tracking* (1 instance): The commercial ILP software solves this instance fast. The commercial LP solver affords good results three times faster. In contrast, dual LP solvers such as ADDD and MPLP do not find good integer solutions.

**Side-Chain Prediction in Protein Folding** (protein-folding). We now consider 21 instances of graphical models for side-chain prediction in protein folding. These instances have many variables and are highly connected. Results are shown in Tab. 21. It can be seen from these results that MPLP, with additional cycle-constraints, obtain the best lower bounds and CombiLP verified optimality for 18 instances, within one hour. The best results are obtained by BPS and LBP followed by lazy flipping with search-depth 2. The rounding techniques used in the algorithms based on linear programming are insufficient for these instances.

**Prediction Protein-Protein Interactions** (protein-prediction). We now consider eight instances of a higher-order graphical model for prediction of protein-protein interactions. Results are shown in Tab. 22. The range of algorithms applicable to these instances is limited. ILP solvers found and verified solutions for 3 instances and performed best for 7 in-stances, within one hour. For one remaining instances, the ILP solver affords a solution far the optimum. Thus, LBP with lazy flipping gives the best results on average.

| algorithm | runtime | value | bound | best | opt |
|---|---|---|---|---|---|
| ogm-LBP-LF2 | 582.65 sec | **−5923.01** | −∞ | 12 | 0 |
| ogm-LF-2 | 87.56 sec | −5747.56 | −∞ | 0 | 0 |
| ogm-TRWS-LF2 | 78.40 sec | −5897.06 | −6041.38 | 6 | 1 |
| BPS-TAB | **25.34** sec | **−5917.15** | −∞ | 12 | 0 |
| ogm-LBP-0.5 | 509.21 sec | −5846.70 | −∞ | 13 | 0 |
| ogm-TRBP-0.5 | 722.27 sec | −5810.68 | −∞ | 9 | 0 |
| ADDD | 251.49 sec | −4213.94 | −10364.89 | 0 | 0 |
| MPLP | 753.55 sec | −5611.60 | −6033.98 | 1 | 1 |
| MPLP-C | 1705.26 sec | −5824.14 | **−5991.75** | 11 | 7 |
| ogm-ADSAL | 1890.32 sec | −5881.78 | −6130.86 | 5 | 1 |
| ogm-BUNDLE-A+ | 1140.61 sec | −5448.03 | −6395.59 | 2 | 1 |
| TRWS-TAB | 25.18 sec | −5771.50 | −6041.38 | 2 | 1 |
| ogm-CombiLP | 568.86 sec | −5911.12 | −6016.78 | **18** | **18** |

Table 21: *protein-folding* (21 instances): Due to the large number of labels, combinatorial methods are not applicable. A notable exception is CombiLP, which manages to solve 18 instances to optimality. MPLP-C gives the best results in terms of the lower bound, but is not the best in generating labelings. The best integer solutions are obtained by BPS and LBP-LF2.

| algorithm | runtime | value | bound | best | opt |
|---|---|---|---|---|---|
| ogm-LBP-LF2 | **69.86** sec | **52942.95** | −∞ | **1** | 0 |
| ogm-LF-3 | 25.99 sec | 57944.06 | −∞ | 0 | 0 |
| ogm-LBP-0.5 | 60.97 sec | 53798.89 | −∞ | 0 | 0 |
| ogm-TRBP-0.5 | 86.03 sec | 61386.17 | −∞ | 0 | 0 |
| ADDD | 10.97 sec | 106216.86 | 41124.16 | 0 | 0 |
| MPLP | 86.60 sec | 101531.75 | 43123.68 | 0 | 0 |
| ogm-LP-LP | 180.72 sec | 102918.41 | 44347.16 | 0 | 0 |
| ogm-ILP | 2262.83 sec | 60047.02 | **44516.07** | **7** | **3** |
| BRAOBB-1 | 3600.16 sec | 61079.07 | −∞ | 0 | 0 |

Table 22: *protein-prediction* (8 instances): Except for one instance, the commercial ILP software solves all instances. LBP followed by lazy flipping is a good approximation.

# 7 Discussion and Conclusions

Our comparative study has shown that there is no single technique which works best for all cases. The reason for this is that, firstly, the models are rather diverse and, secondly, there are several objectives, e.g. running time, best energy or loss functions. We would also like to point out, again, that not all compared models are on the same level of generality and modularity, and some implementations are highly optimized compared to others. Consequently, we do not advise to overemphasize the relative runtimes reported for the different methods which have different implementation paradigms. Moreover, sometimes a single instance can decrease the average performance of a method, which is reported in Tab. 2–22. Consequently, methods that are best for all but one instance are not leading in the average score, e.g. MCI for color-seg-n4 or ILP for protein-prediction.

For most models, we have found that approximative methods provide nearly optimal solutions very fast. According to application specific measurements, these are often not worse than those with optimal energy. With the suggested (heuristic) stopping conditions, TRWS performs well in those case it which it is applicable. This is also due to the fact that for many models the local polytope relaxation is nearly tight. FastPD and $\alpha$-expansion are slightly worse, but have a deterministic stopping condition and do not suffer on problems where the local polytope relaxation is loose, e.g. mrf-photomontage. While methods that can deal with higher order-terms are inferior in terms of runtimes for second order models, they can be directly applied to higher-order models, where best performing methods for second-order models are no longer applicable.

For difficult models like dtf-chinesechar and matching, the exact solutions are significantly better than approximate ones, both in terms of energy and quality. For easier models they can be used to calculate the optimal solutions and allow assessing if approximative methods can be used in applications without harm. Fig. 7 gives an overview of models where all (green), some (yellow) or none (red) instances have been solved to optimality within one hour. As reported in [61] more instances, including those from mrf-inpainting and mrf-photomontage, can be solved to optimality with a time limit of less than a day.

For some models combinatorial optimization methods are *faster* than currently reported state-of-the-art methods. While for small problems combinatorial methods can often be applied directly, for larger problems the reducing of the problem size by partial optimality is required to make them tractable. Solutions from these exact methods are used for evaluating the sub-optimality of the approximate solutions.

Furthermore we observe that the energy-optimal labeling is not always best in terms of application specific loss function, cf. Tab 13. While methods that find global optimal solutions select the global optima – regardless of similar solutions that have a considerable higher energy –, approximative methods often tend to avoid such "isolated" solutions and prefer more "consistent" modes. While this shows that full evaluation of models is only possible in presence of optimal solvers, it also raises the question if approximative methods are preferable when they are not so sensitive to optimal "outliers" or if the models itself need to be improved. The answer to this question might vary for different applications and models and as we have shown, for many models the energy correlates quite well with the loss functions.

Methods based on LP-relaxations over the local polytope often lead to empirically good solutions and are in general not restricted to special subclasses but are also not the fastest ones. Recently, Prua and Werner [59] showed that solving LPs over the local polytope is as hard as solving general LPs, which are of high polynomial complexity. When the local polytope relaxation is too weak, tightening the polytope can help a lot, multicuts for Potts models and MPLP-C for the matching instances are examples of that kind.

While our study covers a broad range of models used nowadays (without being all-embracing), the models used in the last decade might have been biased by solvers that are available and work well. Consequently, second order Potts or truncated convex regularized models, as considered in [66], were in the focus of research. In this study we show alternative methods that can deal with more complex models, including higher order and more densely structured models, cf. dtf-chinesechar, matching or protein-prediction.

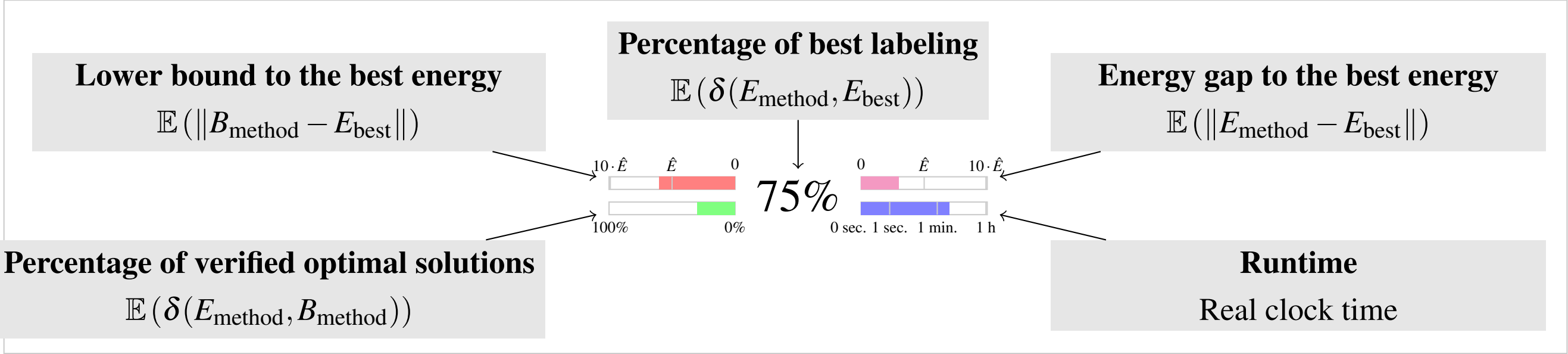

Figure 13: **Legend for the box-plots.** Values are up to numerical precision and averaged over all instances of a model. The bars for energy gap (upper right), gap of the lower bound to the optimal or best available energy (upper left), and runtime (lower right) are scaled piecewise linear between the ticks. The model specific normalization $\hat{E}$,was manually choosen for better visualization and comparability of different datasets.
For numerical reasons we test for a relaxed zero gap using an absolute and relative precision threshold by the test function $\delta(A,B) = \|A-B\| < 10^{-5}$ or $\frac{\|A-B\|}{\|A\|+1} < 10^{-10}$.

With availability of more general optimization methods we hope to stimulate the use of complex and powerful discrete models. This may then inspire the development of new, efficient approximative methods that can meet hard time-constraints in real world applications.

**Acknowledgements.**
We thank Rick Szeliski and Pushmeet Kohli for inspiring discussions. This work has been supported by the German Research Foundation (DFG) within the program “Spatio- / Temporal Graphical Models and Applications in Image Analysis”, grant GRK 1653.

# References

[1] T. Achterberg, T. Koch, and A. Martin. Branching rules revisited. *Operations Research Letters*, 33(1):42 – 54, 2005.

[2] K. Alahari, P. Kohli, and P. H. S. Torr. Reduce, reuse & recycle: Efficiently solving multi-label MRFs. In *In CVPR*, 2008.

[3] K. Alahari, P. Kohli, and P. H. S. Torr. Dynamic hybrid algorithms for MAP inference in discrete MRFs. *IEEE PAMI*, 32(10):1846–1857, 2010.

[4] B. Andres, T. Beier, and J. H. Kappes. OpenGM2. http://hci.iwr.uni-heidelberg.de/opengm2/.

[5] B. Andres, T. Beier, and J. H. Kappes. OpenGM: A C++ library for discrete graphical models. *ArXiv e-prints*, 2012.

[6] B. Andres, J. H. Kappes, T. Beier, U. Köthe, and F. A. Hamprecht. Probabilistic image segmentation with closedness constraints. In *ICCV*, 2011.

[7] B. Andres, J. H. Kappes, T. Beier, U. Köthe, and F. A. Hamprecht. The lazy flipper: Efficient depth-limited exhaustive search in discrete graphical models. In *ECCV*, 2012.

[8] B. Andres, J. H. Kappes, U. Köthe, C. Schnörr, and F. A. Hamprecht. An empirical comparison of inference algorithms for graphical models with higher order factors using OpenGM. In *DAGM*, 2010.

[9] B. Andres, U. Köthe, T. Kroeger, M. Helmstaedter, K. L. Briggman, W. Denk, and F. A. Hamprecht. 3D segmentation of SBFSEM images of neuropil by a graphical model over supervoxel boundaries. *Medical Image Analysis*, 16(4):796–805, 2012.

[10] B. Andres, T. Kröger, K. L. Briggman, W. Denk, N. Korogod, G. Knott, U. Köthe, and F. A. Hamprecht. Globally optimal closed-surface segmentation for connectomics. In *ECCV*, 2012.

[11] D. Batra and P. Kohli. Making the right moves: Guiding alpha-expansion using local primal-dual gaps. In *Computer Vision and Pattern Recognition (CVPR), 2011 IEEE Conference on*, pages 1865–1872. IEEE, 2011.

[12] M. Bergtholdt, J. H. Kappes, S. Schmidt, and C. Schnörr. A study of parts-based object class detection using complete graphs. *IJCV*, 87(1-2):93–117, 2010.

[13] J. Besag. On the Statistical Analysis of Dirty Pictures. *Journal of the Royal Statistical Society. Series B (Methodological)*, 48(3):259–302, 1986.

[14] Y. Boykov. Computing geodesics and minimal surfaces via graph cuts. In *ICCV*, 2003.

[15] Y. Boykov, O. Veksler, and R. Zabih. Fast approximate energy minimization via graph cuts. *IEEE PAMI*, 23(11):1222–1239, 2001.

[16] U. Brandes, D. Delling, M. Gaertler, R. Görke, M. Hoefer, Z. Nikoloski, and D. Wagner. On modularity clustering. *IEEE Transactions on Knowledge and Data Engineering*, 20(2):172–188, 2008.

[17] C. A. Cocosco, V. Kollokian, R.-S. Kwan, and A. C. Evans. Brainweb: Online interface to a 3d mri simulated brain database. *NeuroImage*, 5(4), May 1997.

[18] IBM ILOG CPLEX Optimizer. `http://www-01.ibm.com/software/integration/optimization/cplex-optimizer/`, 2013.

[19] A. Delong, A. Osokin, H. Isack, and Y. Boykov. Fast approximate energy minimization with label costs. *International Journal of Computer Vision*, 96:1–27, Jan. 2012.

[20] G. Elidan and A. Globerson. The probabilistic inference challenge (PIC2011). http://www.cs.huji.ac.il/project/PASCAL/.

[21] P. F. Felzenszwalb and D. P. Huttenlocher. Efficient belief propagation for early vision. *Int. J. Comput. Vision*, 70(1):41–54, Oct. 2006.

[22] A. Fix, A. Gruber, E. Boros, and R. Zabih. A graph cut algorithm for higher-order Markov random fields. In *ICCV*, 2011.

[23] A. C. Gallagher, D. Batra, and D. Parikh. Inference for order reduction in Markov random fields. In *CVPR*, 2011.

[24] A. Globerson and T. Jaakkola. Fixing max-product: Convergent message passing algorithms for MAP LP-relaxations. In *NIPS*, 2007.

[25] D. Goldberg. What every computer scientist should know about floating-point arithmetic. *ACM Comput. Surv.*, 23(1):5–48, 1991.

[26] S. Gould, R. Fulton, and D. Koller. Decomposing a scene into geometric and semantically consistent regions. In *ICCV*, 2009.

[27] M. Guignard and S. Kim. Lagrangean decomposition: a model yielding stronger lagrangean bounds. *Mathematical programming*, 39(2):215–228, 1987.

[28] D. Hoiem, A. A. Efros, and M. Hebert. Recovering occlusion boundaries from an image. *IJCV*, 91(3):328–346, 2011.

[29] F. Hutter, H. H. Hoos, and T. Stützle. Efficient stochastic local search for MPE solving. In L. P. Kaelbling and A. Saffiotti, editors, *IJCAI*, pages 169–174, 2005.

[30] A. Jaimovich, G. Elidan, H. Margalit, and N. Friedman. Towards an integrated protein-protein interaction network: A relational markov network approach. *Journal of Computational Biology*, 13(2):145–164, 2006.

[31] J. H. Kappes, B. Andres, F. A. Hamprecht, C. Schnörr, S. Nowozin, D. Batra, S. Kim, B. X. Kausler, J. Lellmann, N. Komodakis, and C. Rother. A Comparative Study of Modern Inference Techniques for Discrete Energy Minimization Problem. In *CVPR*, 2013.

[32] J. H. Kappes, B. Savchynskyy, and C. Schnörr. A bundle approach to efficient MAP-inference by Lagrangian relaxation. In *CVPR*, 2012.

[33] J. H. Kappes, M. Speth, B. Andres, G. Reinelt, and C. Schnörr. Globally optimal image partitioning by multicuts. In *EMMCVPR*, 2011.

[34] J. H. Kappes, M. Speth, G. Reinelt, and C. Schnörr. Higher-order segmentation via multicuts. *ArXiv e-prints*, May 2013. http://arxiv.org/abs/1305.6387.

[35] J. H. Kappes, M. Speth, G. Reinelt, and C. Schnörr. Towards efficient and exact MAP-inference for large scale discrete computer vision problems via combinatorial optimization. In *CVPR*, 2013.

[36] B. X. Kausler, M. Schiegg, B. Andres, M. Lindner, H. Leitte, L. Hufnagel, U. Koethe, and F. A. Hamprecht. A discrete chain graph model for 3d+t cell tracking with high misdetection robustness. In *ECCV*, 2012.

[37] B. W. Kernighan and S. Lin. An efficient heuristic procedure for partitioning graphs. *The Bell Systems Technical Journal*, 49(2):291–307, 1970.

[38] S. Kim, S. Nowozin, P. Kohli, and C. D. Yoo. Higher-order correlation clustering for image segmentation. In *NIPS*, pages 1530–1538, 2011.

[39] T. Kim, S. Nowozin, P. Kohli, and C. D. Yoo. Variable grouping for energy minimization. In *CVPR*, pages 1913–1920, 2011.

[40] P. Kohli, L. Ladicky, and P. Torr. Robust higher order potentials for enforcing label consistency. *International Journal of Computer Vision*, 82(3):302–324, 2009.

[41] D. Koller and N. Friedman. *Probabilistic Graphical Models: Principles and Techniques*. MIT Press, 2009.

[42] V. Kolmogorov. Convergent tree-reweighted message passing for energy minimization. *PAMI*, 28(10):1568–1583, 2006.

[43] V. Kolmogorov and C. Rother. Comparison of energy minimization algorithms for highly connected graphs. In *ECCV*, pages 1–15, 2006.

[44] V. Kolmogorov and R. Zabih. What energy functions can be minimized via graph cuts? In *ECCV*, 2002.

[45] N. Komodakis and N. Paragios. Beyond loose LP-relaxations: Optimizing MRFs by repairing cycles. In *ECCV*, 2008.

[46] N. Komodakis, N. Paragios, and G. Tziritas. MRF energy minimization and beyond via dual decomposition. *IEEE Transactions on Pattern Analysis and Machine Intelligence*, 33(3):531–552, 2011.

[47] N. Komodakis and G. Tziritas. Approximate labeling via graph cuts based on linear programming. *IEEE PAMI*, 29(8):1436–1453, 2007.

[48] I. Kovtun. Partial optimal labeling search for a np-hard subclass of (max, +) problems. In B. Michaelis and G. Krell, editors, *DAGM-Symposium*, volume 2781 of *Lecture Notes in Computer Science*, pages 402–409. Springer, 2003.

[49] J. Lellmann and C. Schnörr. Continuous Multiclass Labeling Approaches and Algorithms. *SIAM J. Imag. Sci.*, 4(4):1049–1096, 2011.

[50] V. Lempitsky, C. Rother, S. Roth, and A. Blake. Fusion moves for markov random field optimization. *IEEE Transactions on Pattern Analysis and Machine Intelligence*, 32(8):1392–1405, 2010.

[51] A. F. T. Martins, M. A. T. Figueiredo, P. M. Q. Aguiar, N. A. Smith, and E. P. Xing. An augmented lagrangian approach to constrained MAP inference. In *ICML*, pages 169–176, 2011.

[52] C. Nieuwenhuis, E. Toeppe, and D. Cremers. A survey and comparison of discrete and continuous multi-label optimization approaches for the Potts model. *International Journal of Computer Vision*, 2013.

[53] S. Nowozin and C. H. Lampert. Structured learning and prediction in computer vision. *Foundations and Trends in Computer Graphics and Vision*, 6(3–4):185–365, 2011.

[54] S. Nowozin, C. Rother, S. Bagon, T. Sharp, B. Yao, and P. Kohli. Decision tree fields. In *ICCV*, pages 1668–1675. IEEE, 2011.

[55] F. Orabona, T. Hazan, A. Sarwate, and T. Jaakkola. On Measure Concentration of Random Maximum A-Posteriori Perturbations. In *Proc. ICML*, 2014.

[56] A. Osokin, D. Vetrov, and V. Kolmogorov. Submodular decomposition framework for inference in associative markov networks with global constraints. In *CVPR*, pages 1889–1896, 2011.

[57] L. Otten and R. Dechter. Anytime AND/OR depth-first search for combinatorial optimization. In *Proceedings of the Annual Symposium on Combinatorial Search (SOCS)*, 2011.

[58] G. Papandreou and A. Yuille. Perturb-and-MAP Random Fields: Using Discrete Optimization to Learn and Sample from Energy Models. In *Proc. ICCV*, 2011.

[59] D. Prua and T. Werner. Universality of the local marginal polytope. In *CVPR*, pages 1738–1743. IEEE, 2013.

[60] C. Rother, V. Kolmogorov, V. S. Lempitsky, and M. Szummer. Optimizing binary MRFs via extended roof duality. In *CVPR*, 2007.

[61] B. Savchynskyy, J. H. Kappes, P. Swoboda, and C. Schnörr. Global MAP-optimality by shrinking the combinatorial search area with convex relaxation. In *NIPS*, 2013.

[62] B. Savchynskyy and S. Schmidt. Getting feasible variable estimates from infeasible ones: MRF local polytope study. In *Workshop on Inference for Probabilistic Graphical Models at ICCV 2013*, 2013.

[63] B. Savchynskyy, S. Schmidt, J. H. Kappes, and C. Schnörr. Efficient MRF energy minimization via adaptive diminishing smoothing. In *UAI 2012*, pages 746–755, 2012.

[64] M. Schlesinger. Sintaksicheskiy analiz dvumernykh zritelnikh signalov v usloviyakh pomekh (Syntactic analysis of two-dimensional visual signals in noisy conditions). *Kibernetika*, 4:113–130, 1976.

[65] D. Sontag, D. K. Choe, and Y. Li. Efficiently searching for frustrated cycles in MAP inference. In N. de Freitas and K. P. Murphy, editors, *UAI*, pages 795–804. AUAI Press, 2012.

[66] R. Szeliski, R. Zabih, D. Scharstein, O. Veksler, V. Kolmogorov, A. Agarwala, M. Tappen, and C. Rother. A comparative study of energy minimization methods for Markov random fields with smoothness-based priors. *IEEE PAMI*, 30(6):1068–1080, 2008.

[67] D. Tarlow, D. Batra, P. Kohli, and V. Kolmogorov. Dynamic tree block coordinate ascent. In *Proceedings of the International Conference on Machine Learning (ICML)*, 2011.

[68] T. Verma and D. Batra. Maxflow revisited: An empirical comparison of maxflow algorithms for dense vision problems. In *BMVC*, pages 1–12, 2012.

[69] M. J. Wainwright, T. Jaakkola, and A. S. Willsky. MAP estimation via agreement on trees: message-passing and linear programming. *IEEE Trans. Inf. Theory*, 51(11):3697–3717, 2005.

[70] T. Werner. A linear programming approach to max-sum problem: A review. *IEEE Trans. Pattern Anal. Mach. Intell.*, 29(7):1165–1179, July 2007.

[71] F. Wesselmann and U. Stuhl. Implementing cutting plane management and selection techniques. Technical report, University of Paderborn, 2012. `http://www.optimization-online.org/DB_HTML/2012/12/3714.html`.

[72] C. Yanover, O. Schueler-Furman, and Y. Weiss. Minimizing and learning energy functions for side-chain prediction. *Journal of Computational Biology*, 15(7):899–911, 2008.

[73] J. S. Yedidia, W. T. Freeman, and Y. Weiss. Constructing free energy approximations and generalized belief propagation algorithms. MERL Technical Report, 2004-040, 2004. http://www.merl.com/papers/docs/TR2004-040.pdf.

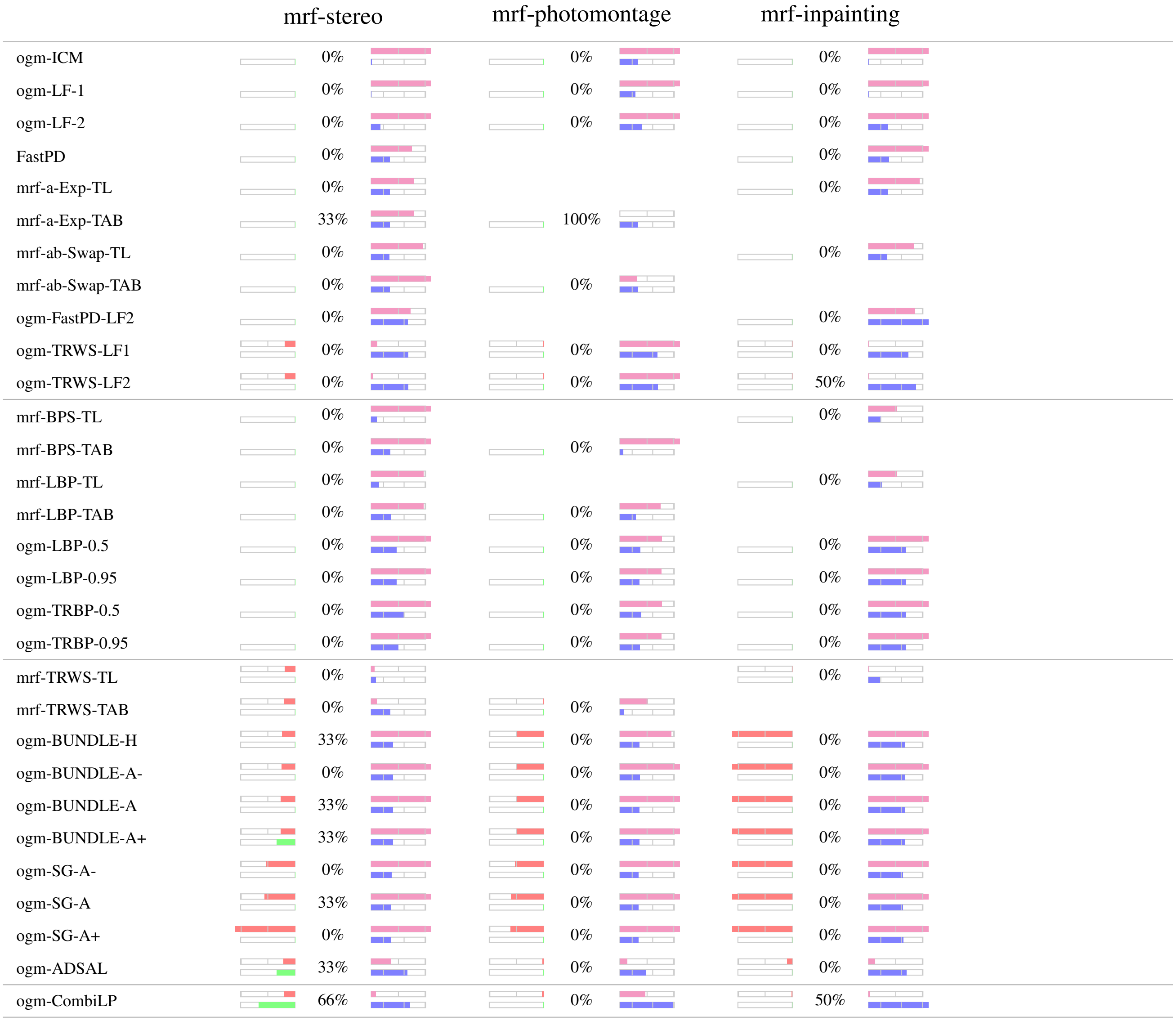

Figure 14: **Results for grid structured models with four-pixel neighborhood-system and truncated convex regularization** from [66]. See Fig. 13 for the legend. For all models local polytope relaxations give good lower bounds. However, extracting a integer solution from the fractional one can be very difficult, especially for the photomontage instances, where soft constraints renders linear programming relaxation harder. FastPD and $\alpha$-expansion are the first choice if fast optimization is required. For stereo- and inpainting problems LP-relaxations as obtained by TRWS have given best results – especially when followed by lazy flipping post processing. For some instances we were able to solve them to optimality by CombiLP but often need more than 1 hour.

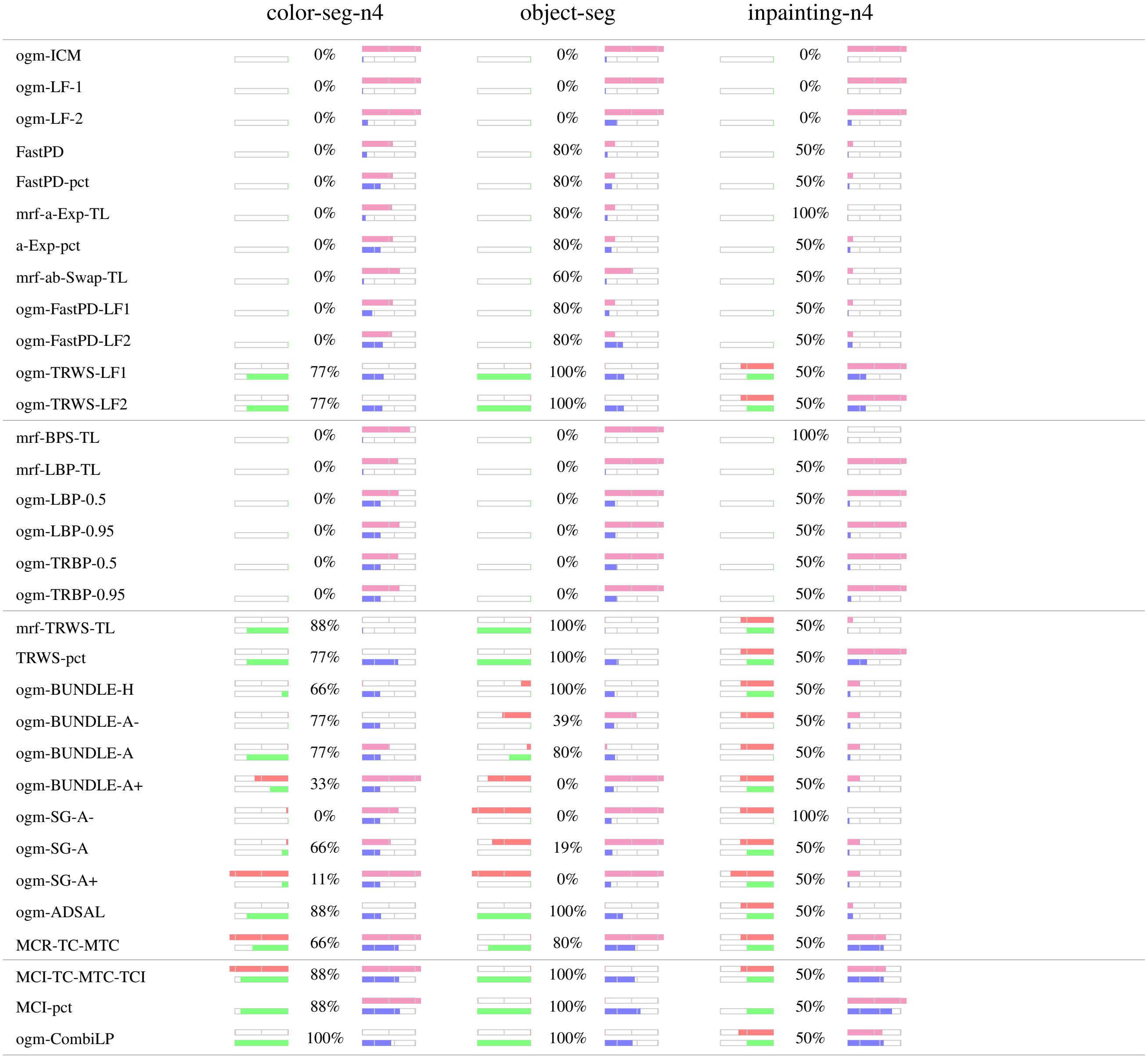

Figure 15: **Results for grid structured models with four-pixel neighborhood-system and Potts regularization.** See Fig. 13 for the legend. For Potts models relaxations over the local polytope are often very tight and multicut relaxations works quite good and efficient. FastPD is the first choice if fast optimization is required. When ever persistence is available to reduce the problem size it helps a lot. In the color-seg-n4 dataset the instance pfau is the hardest one and rises the average runtime. Also the lower bound on the pfau instances produced by MCI within one hour are very weak. The fastest exact solver is here CombiLP. In the inpainting-n4 dataset the inverse instance is designed to be hard and make LP-solvers struggling.

| | color-seg-n8 | color-seg | inpainting-n8 |
|---|---|---|---|
| ogm-ICM | 0% | 0% | 0% |
| ogm-LF-1 | 0% | 0% | 0% |
| ogm-LF-2 | 0% | 0% | 0% |
| FastPD | 0% | 66% | 50% |
| FastPD-pct | 0% | 66% | 100% |
| a-Exp-VIEW | 0% | 66% | 50% |
| a-Exp-pct | 0% | 100% | 100% |
| ab-Swap-VIEW | 0% | 66% | 50% |
| ogm-FastPD-LF1 | 0% | 66% | 50% |
| ogm-FastPD-LF2 | 0% | 66% | 50% |
| ogm-TRWS-LF1 | 33% | 66% | 50% |
| ogm-TRWS-LF2 | 44% | 66% | 50% |
| BPS-TL | 0% | 0% | 50% |
| ogm-LBP-0.5 | 0% | 0% | 50% |
| ogm-LBP-0.95 | 0% | 0% | 50% |
| ogm-TRBP-0.5 | 0% | 0% | 50% |
| ogm-TRBP-0.95 | 0% | 0% | 50% |
| TRWS-TL | 22% | 66% | 50% |
| TRWS-pct | 11% | 66% | 50% |
| ogm-BUNDLE-H | 33% | 33% | 0% |
| ogm-BUNDLE-A- | 22% | 33% | 50% |
| ogm-BUNDLE-A | 44% | 33% | 50% |
| ogm-BUNDLE-A+ | 11% | 33% | 50% |
| ogm-SG-A- | 0% | 0% | 0% |
| ogm-SG-A | 11% | 0% | 0% |
| ogm-SG-A+ | 0% | 0% | 0% |
| ogm-ADSAL | 55% | 100% | 50% |
| MCR-TC-MTC | 33% | 100% | 50% |
| ogm-CombiLP | 100% | 100% | 50% |
| MCI-TC-MTC-TCI | 88% | 100% | 50% |
| MCI-TC-TCI | | 100% | 50% |
| MCI-pct | 77% | 100% | 50% |

Figure 16: **Results for grid structured models with 8-pixel neighborhood-system and Potts regularization.** See Fig. 13 for the legend. For Potts models relaxations over the local polytope are often very tight and multicut relaxations works quite good and efficient. Compare to the same models with a 4-pixel neighborhood-system, cf. Fig. 15, the local polytope relaxations becomes weaker. FastPD is the first choice if fast optimization is required. When ever persistence is available to reduce the problem size it helps a lot. In the color-seg-n8 dataset the instance pfau is the hardest one and rises the average runtime. Also the lower bound on the pfau instances produced by MCA within 6 hours are very weak. In the inpainting-n8 dataset the inverse instance is designed to be hard and make LP-solvers struggling.

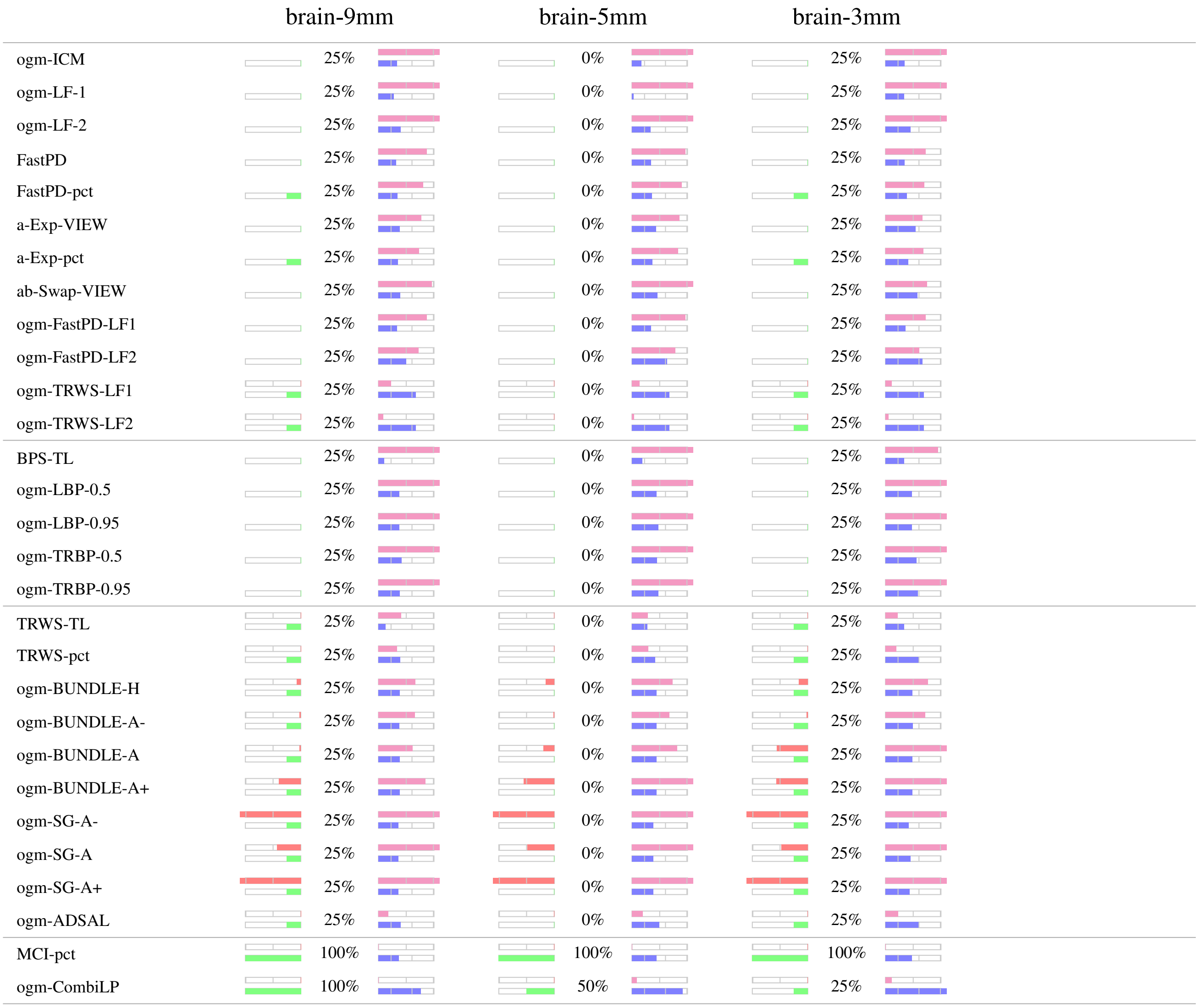

Figure 17: **Results for grid structured models with 3D 6-voxel neighborhood-system and Potts regularization.** See Fig. 13 for the legend. The slice thick is given in millimeters. Thinner slices give larger models. FastPD is the fastest and TRWS the best approximative method. Local polytope relaxations give very good bounds for this models. Exact results are feasible by MCA-pct with runtimes comparable to approximative methods. Without preprocessing (pct) MCA is much slower and requires more than 10 GB of memory for the smallest instance.

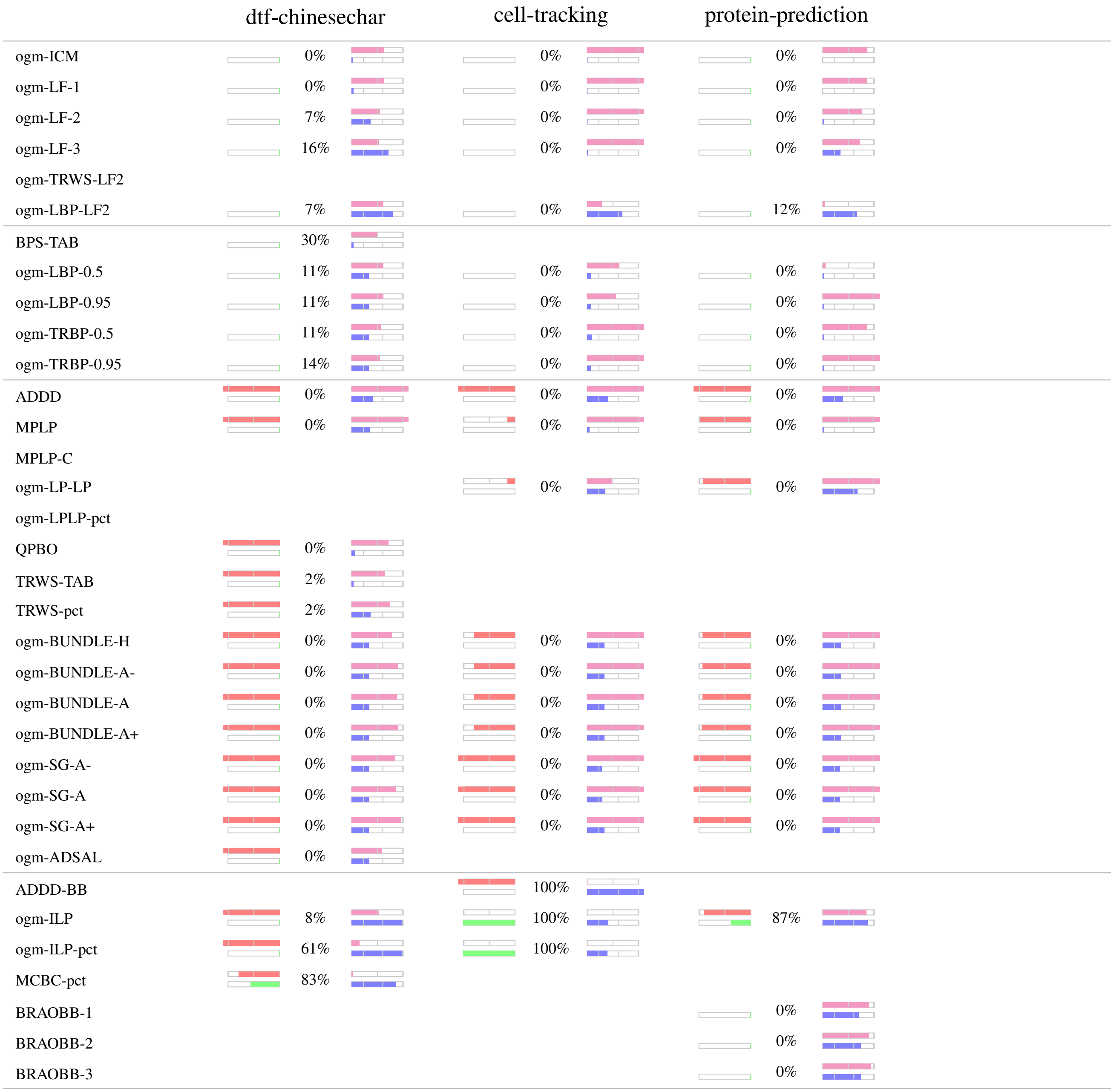

Figure 18: **Results for binary models.** See Fig. 13 for the legend. The Chinese character instances include Potts terms with negative coupling strength, that means the models are non-submodular. That why only a small subset of solvers is applicable. The large degree of the structure makes local polytope relaxations week and the remaining rounding problem hard. BPS give good results, but we obtain best results by first using persistence to reduce the problem size and then tightening the relaxation. The cell-tracking and protein-prediction dataset include higher order terms. This render them more challenging and many standard methods are no longer applicable. For cell-tracking ogm-ILP performs best, many other methods violate soft-constraints which causes hight objective values. For the protein-protein dataset ogm-ILP performs best except of 1 instance. This seemed to be caused by numerical problems. For all protein-protein instances LBP, optionally followed by lazy flipping give very good results in reasonable time.

| | inclusion | geo-surf-3 | geo-surf-7 |
|---|---|---|---|
| A-Fusion | 0% | 97% | 85% |
| ogm-ICM | 0% | 44% | 2% |
| ogm-LF-1 | 0% | 44% | 2% |
| ogm-LF-2 | 0% | 69% | 4% |
| ogm-LF-3 | 0% | 87% | 8% |
| ogm-LBP-LF2 | 69% | 91% | 22% |
| ogm-LBP-0.5 | 69% | 90% | 22% |
| ogm-LBP-0.95 | 80% | 90% | 22% |
| ogm-TRBP-0.5 | 50% | 94% | 42% |
| ogm-TRBP-0.95 | 80% | 94% | 39% |
| ADDD | 10% | 97% | 98% |
| MPLP | 19% | 97% | 92% |
| MPLP-C | 19% | 98% | 94% |
| ogm-LP-LP | 10% | 100% | 99% |
| ogm-BUNDLE-H | 30% | 100% | 99% |
| ogm-BUNDLE-A- | 39% | 99% | 36% |
| ogm-BUNDLE-A | 10% | 100% | 86% |
| ogm-BUNDLE-A+ | 0% | 100% | 99% |
| ogm-SG-A- | 0% | 89% | 14% |
| ogm-SG-A | 0% | 100% | 53% |
| ogm-SG-A+ | 0% | 100% | 98% |
| ogm-ILP | 100% | 100% | 100% |
| BRAOBB-1 | 0% | 99% | 71% |
| BRAOBB-2 | 0% | 99% | 88% |
| BRAOBB-3 | 0% | 100% | 89% |

Figure 19: **Results for higher order models.** See Fig. 13 for the legend. For the inclusion instances ogm-ILP give best results, similar to those of LBP. LP relaxations of the local polytope are relatively tight but rounding is not trivial and often violates the inclusion soft-constraints. Adding additional cycle constraints does improve the results only marginally. The geo-surf instances are based on superpixels and therefor much smaller. Fastest and optimal results are produced by ogm-ILP and AD3-BB. Later is non-commercial available under the LGPL.

| | correlation-clustering | image-seg | image-seg-3rdorder | modularity-clustering |
|---|---|---|---|---|
| ogm-KL | | 0% | | 50% |
| ogm-ICM | 0% | 0% | 0% | 0% |
| ogm-LF-1 | 0% | 0% | 0% | 0% |
| MCR-CC | 22% | 36% | 0% | 16% |
| MCR-CCFDB | 22% | 35% | 0% | 16% |
| MCR-CCFDB-OWC | 22% | 35% | 0% | 83% |
| MCI-CCI | 100% | 100% | 99% | 66% |
| MCI-CCIFD | 100% | 100% | 99% | 66% |
| MCI-CCFDB-CCIFD | 100% | 100% | 100% | 83% |

Figure 20: **Results for unsupervised partition models.** See Fig. 13 for the legend. Over all MCI-CCFDB-CCIFD performs best. The KL-method used outside computer vision does not perform well on sparse computer vision problems and cannot be used for higher order models. Linear programming relaxations give worse results and are not necessary faster than ILP-based methods. The reason for this is, that the separation procedures for non-integer solutions are more complex and time consuming.

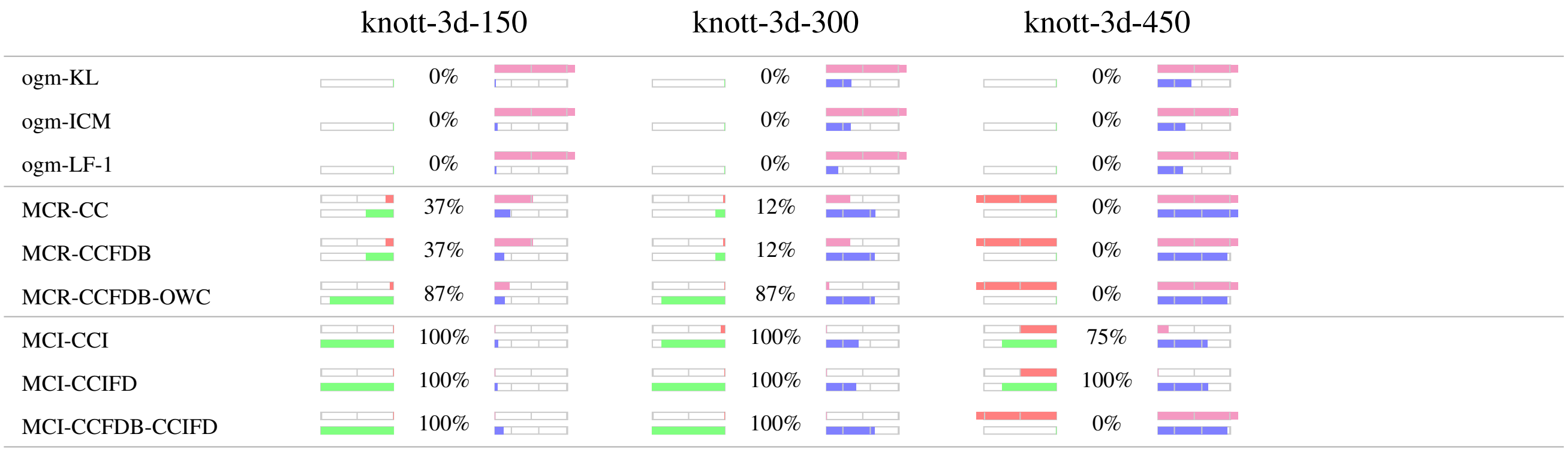

Figure 21: **Evaluation of the 3D neuron segmentation datasets.** See Fig. 13 for the legend. With increasing problem size relaxations (MCR) get worse. Also integer variants (MCI) suffers and separating violated constraints becomes the most time consuming part. Consequently, for large models it is preferable to start directly with integer constraints and not to start with an LP-relaxation first, as done within MCI-CCFDB-CCIFD, because the separation procedure is than to slow.

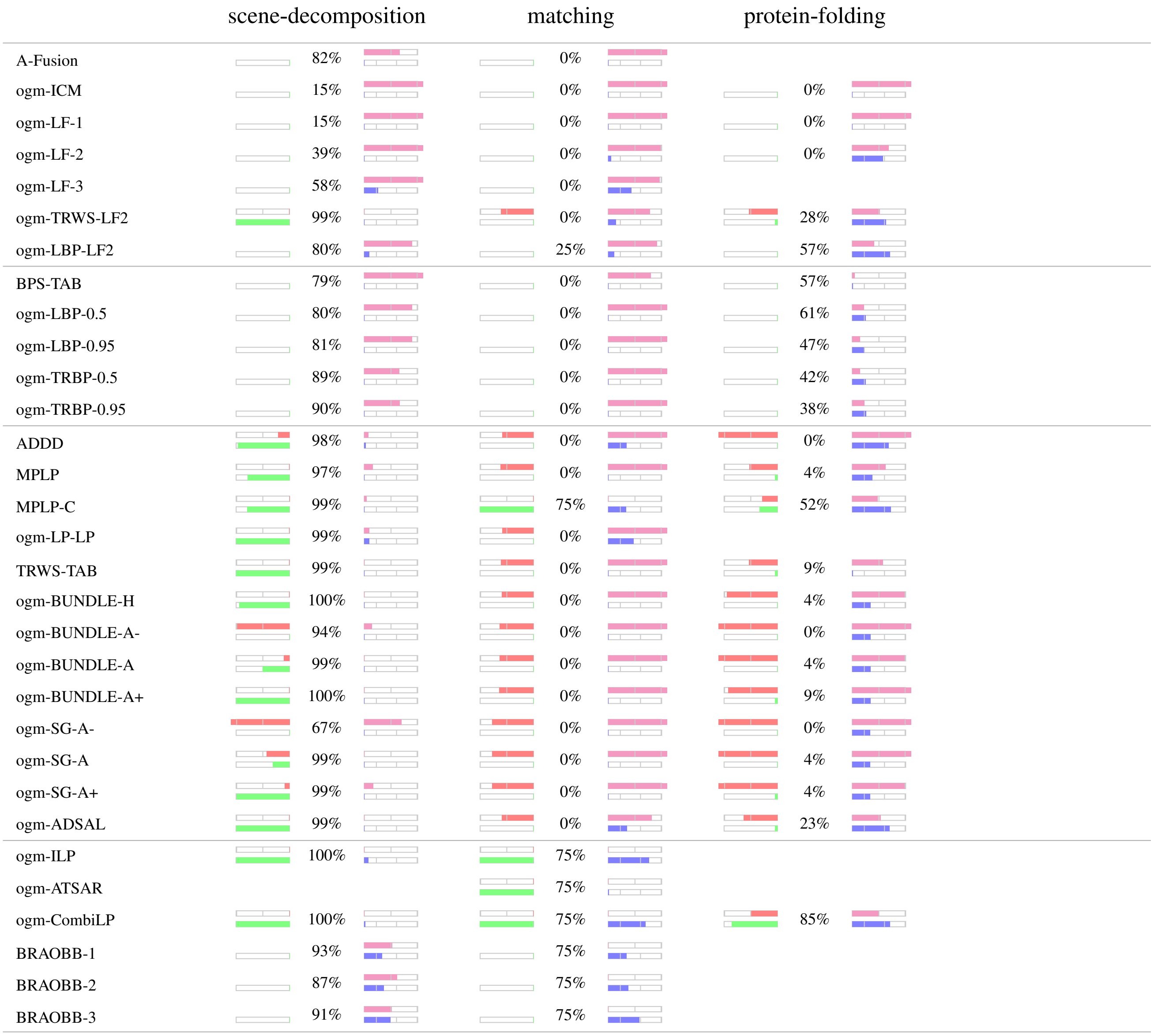

Figure 22: **Evaluation of the second order models with no truncated convex regularizers.** See Fig. 13 for the legend. The scene decomposition instances are based on superpixels, such models are small and combinatorial methods like ogm-ILP or AD3-BB are fast and optimal. Contrary to scene-decomposition for the matching instances the local polytope relaxation is not tight. On can either tighten the relaxation MPLP-C or use alternative methods to obtain bounds ogm-AStar for obtaining fast optimal results. For protein-folding relaxations are weak too and the huge label-space renders the problem hard for many solvers, e.g. ogm-ILP. We obtain the best results by BPS.